\pdfoutput=1

\documentclass[11pt]{article}

\usepackage[preprint]{acl}

\usepackage{times}
\usepackage{latexsym}

\usepackage[T1]{fontenc}

\usepackage[utf8]{inputenc}

\usepackage{microtype}

\usepackage{inconsolata}

\usepackage{graphicx}
\usepackage{multirow}
\usepackage{makecell}
\usepackage{booktabs}
\usepackage{verbatim}

\title{DentalBench: Benchmarking and Advancing LLMs Capability for \\   Bilingual Dentistry Understanding}

\author{
  Hengchuan Zhu\textsuperscript{1}, Yihuan Xu\textsuperscript{1}, Yichen Li\textsuperscript{1}, Zijie Meng\textsuperscript{1}, Zuozhu Liu\textsuperscript{1, 2}\footnotemark[1] \\
  \textsuperscript{1}Zhejiang University \\
  \textsuperscript{2}ZJU-Angelalign R\&D Center for Intelligence Healthcare, Zhejiang, China \\
  \texttt{\{zhuhengchuan1.24, zuozhuliu\}@intl.zju.edu.cn}
}

\begin{document}
\maketitle
\footnotetext[1]{Corresponding author.}
\begin{abstract}
Recent advances in large language models (LLMs) and medical LLMs (Med-LLMs) have demonstrated strong performance on general medical benchmarks. However, their capabilities in specialized medical fields, such as dentistry which require deeper domain-specific knowledge, remain underexplored due to the lack of targeted evaluation resources. In this paper,  we introduce \textbf{DentalBench}, the first comprehensive bilingual benchmark designed to evaluate and advance LLMs in the dental domain. DentalBench consists of two main components: \textbf{DentalQA}, an English-Chinese question-answering (QA) benchmark with 36,597 questions spanning 4 tasks and 16 dental subfields; and \textbf{DentalCorpus}, a large-scale, high-quality corpus with 337.35 million tokens curated for dental domain adaptation, supporting both supervised fine-tuning (SFT) and retrieval-augmented generation (RAG). We evaluate 14 LLMs, covering proprietary, open-source, and medical-specific models, and reveal significant performance gaps across task types and languages. Further experiments with Qwen-2.5-3B demonstrate that domain adaptation substantially improves model performance, particularly on knowledge-intensive and terminology-focused tasks, and highlight the importance of domain-specific benchmarks for developing trustworthy and effective LLMs tailored to healthcare applications.
\end{abstract}




\begin{figure*}
    \centering
    \includegraphics[width=1\linewidth]{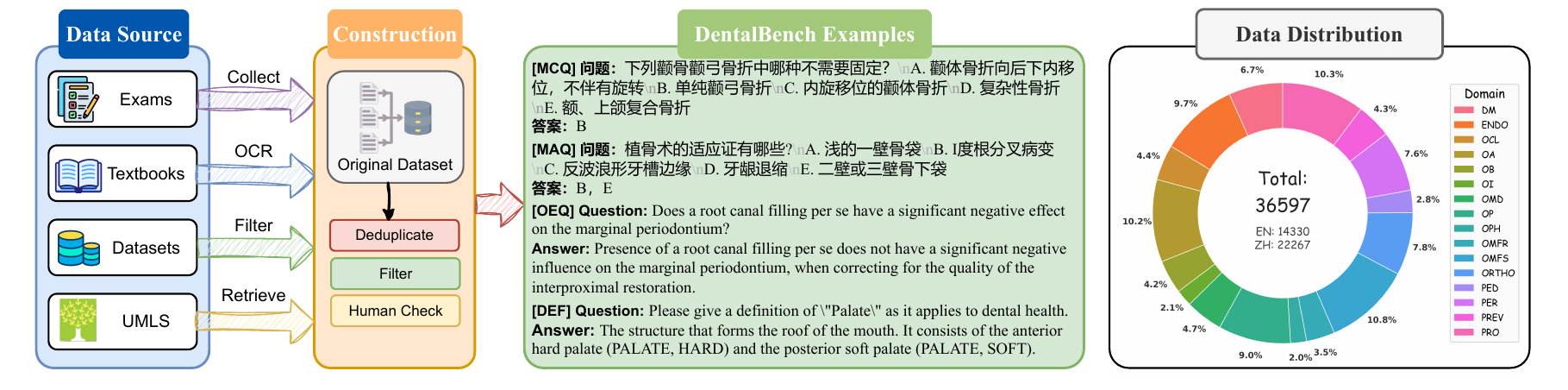}
    \caption{Overview of the DentalBench. It encompasses the following 16 dental specialties and disciplines: dental materials (DM), endodontics (ENDO), occlusion (OCL), oral anatomy (OA), oral biology (OB), oral implantology (OI), oral mucosal diseases (OMD), oral pathology (OP), oral pharmacology (OPH), oral and maxillofacial radiology (OMFR), oral and maxillofacial surgery (OMFS), orthodontics (ORTHO), pediatric dentistry (PED), periodontics (PER), preventive dentistry (PREV), and prosthodontics (PRO).}
    \label{fig:overview}
    \vspace{-1em}
\end{figure*}

\section{Introduction}
Large Language Models (LLMs) have demonstrated impressive capabilities in a wide range of domains~\cite{openai2024gpt4technicalreport,yang2025qwen3,liu2024deepseek,team2025gemma,deepseekai2025deepseekr1incentivizingreasoningcapability, openaio1}. Especially in the medical field, recent studies have shown that LLMs can achieve expert-level performance on various clinical benchmarks~\cite{wu2025towards,llavamed,pmcllama,retfound}. 
However, reliable and fine-grained evaluation of LLM performance in specialized medical subfields-such as dentistry-remain limited, because of the shortage of domain-specific knowledge in general medical corpora or benchmarks. 

As an important and highly specialized branch of medicine that spans multiple subfields and involves complex procedures, oral healthcare is in great need of artificial intelligence integration. Although there have been some studies exploring the integration of deep learning techniques into dentistry~\cite{shi2024leta,wei2020tanet,xiong2023tsegformer,liu2023deep}, LLMs remain under-evaluated due to the lack of targeted evaluation resources. It hinders not only the understanding of current LLM limitations but also the development of robust systems for clinical applications. 


Therefore, in this paper, we introduce \textbf{DentalBench}, a comprehensive benchmark and corpus designed for evaluating and advancing LLM performance in the dental domain. We first construct \textbf{DentalQA}, an English-Chinese question-answering (QA) benchmark covering 4 task formats and 16 specialized subfields. Then, we develop \textbf{DentalCorpus}, a professionally curated bilingual corpus with large-scale and high-quality, aimed at dental-domain adaptation. Using DentalQA, we systematically evaluate various proprietary, open-source and medical-specific LLMs and reveal significant limitations for current models to finish knowledge-intensive tasks in dentistry. Further experiments based on supervised fine-tuning (SFT) and retrieval-augmented generation (RAG) by the DentalCorpus demonstrate that access to in-domain data can substantially improve model performance in specialized oral healthcare tasks,  highlighting the importance of benchmarks for domain adaptation in real-world applications. Our main contributions are summarized as follows:

\begin{itemize}
\setlength{\itemsep}{0pt} 
\setlength{\topsep}{0pt}  
    \item We introduce \textbf{DentalQA}, the first bilingual benchmark for dentistry-specific language understanding, consisting of 36,597 questions across 4 task types and 16 subfields.
    \item We create \textbf{DentalCorpus}, a large-scale, high-quality corpus containing 337.35 million tokens curated for dental domain adaptation with SFT and RAG methods.
    \item We evaluate 14 LLMs—including proprietary, open-source, and medical-specific models—on DentalQA, revealing clear performance gaps across task types and languages. Through extensive experiments, we further demonstrate that domain adaptation with DentalCorpus significantly improves general LLM performance in the dental domain.
\end{itemize}

\section{DentalBench Dataset}

We introduce \textbf{DentalBench}, the first comprehensive dataset for evaluating and adapting LLMs in the dental domain, as shown in Figure~\ref{fig:overview}. It consists of: \textbf{DentalQA}, a bilingual benchmark for evaluating knowledge-based reasoning in oral heathcare, and \textbf{DentalCorpus}, a large-scale and high-quality text corpus curated for dental domain adaptation.

\subsection{DentalQA}
We construct \textbf{DentalQA}, a high-quality English-Chinese benchmark comprising 36,597 questions, covering 4 task formats and 16 dental subfields. 

\noindent\textbf{Task Formats.} DentalQA includes the following four question types:
\textbf{(a) MCQ:} Single-answer multiple choice questions (4 in English, 5 options in Chinese), testing factual recall.
\textbf{(b) MAQ:} Multi-answer multiple choice questions (Chinese only), assessing comprehensive diagnostic knowledge.
\textbf{(c) OEQ:} Open-ended questions simulating clinical and theoretical scenarios, used to evaluate reasoning and generation.
\textbf{(d) DEF:} Terminology definition questions, requiring understanding of domain-specific dental terms.

\noindent\textbf{Domain Coverage.} Each question is categorized into one of 16 dental subfields (e.g., oral anatomy, periodontics, orthodontics) based on standard textbook classifications of the 8th round of the National Higher Education Curriculum for Five-Year Undergraduate Dental Medicine Programs (e.g., \citet{zhao2020orthodontics}). Figure~\ref{fig:overview} shows examples across task formats and domain data distributions, with additional details provided in Appendix~\ref{sec:distribution}.

\noindent\textbf{Data Sources.}  
The English dataset is curated from seven public medical QA datasets: MMLU \cite{hendrycks2021measuringmassivemultitasklanguage}, MedQA \cite{jin2020diseasedoespatienthave}, MedMCQA \cite{pal2022medmcqalargescalemultisubject}, MedQuAD \cite{BenAbacha-BMC-2019}, PubMedQA \cite{jin2019pubmedqadatasetbiomedicalresearch}, and iCliniq \cite{regin2017medical}, Medical Meadow Flashcards and Medical Meadow Wikidoc \cite{yu2024enhancinghealthcarelargelanguage}. Then, we use a keyword list derived from the DentalCorpus filtering process to filter the datasets. Furthermore, we use dental terms from a bilingual glossary compiled from textbooks in the DentalCorpus pipeline and retrieve their definitions from UMLS \cite{umls_nlm} to construct English DEF questions.
The Chinese dataset includes questions from the China National Dental Licensing Examination (1999–2021), 34 dental textbooks and auxiliary materials, and 181 OEQs derived from real orthodontist-patient interactions.  

\noindent\textbf{Construction.}  
We apply a unified pipeline across both languages. MCQs and MAQs are normalized to fixed option counts. DEF questions are generated by filling 50 predefined templates per language (Appendix~\ref{sec:def_templates}) with extracted dental terms and their definitions. OEQs are preserved in their original form without modification.  
To ensure quality and domain relevance, we use GPT-4o to classify all questions into three categories: \textit{oral-related}, \textit{non-oral}, and \textit{insufficient} (Appendix~\ref{sec:classification_prompt}). 
The \textit{insufficient} category is used for questions with incomplete or corrupted content. We filter and retain only the \textit{oral-related} questions.

\noindent\textbf{Human Validation.}
To assess classification accuracy, we manually reviewed 300 representative samples—50 for each combination of language and category. The results indicate strong agreement: 100\%, 96\%, and 94\% for English, and 96\%, 92\%, and 92\% for Chinese.

\subsection{DentalCorpus}

We construct \textbf{DentalCorpus}, a bilingual resource designed to support domain adaptation and retrieval-augmented generation in dentistry. 


\noindent\textbf{Data Sources.} DentalCorpus is built from three major sources:
\textbf{(a) Textbooks.} We collect 40 Chinese dental textbooks and auxiliary materials, remove non-content sections and apply OCR to obtain 4.1M characters of clean text. We also extract a bilingual glossary of 1,971 dental terms from glossaries.
\textbf{(b) PubMed Articles.} Using 28 MeSH terms (listed in Appendix~\ref{sec:mesh_terms}), we retrieve 54,651 freely accessible full-text articles from PubMed~\cite{pubmed}, published between 2000 and 2024, yielding 983.3M English and 5.4M Chinese characters.
\textbf{(c) Open Medical Datasets.} We filter MMedC \cite{qiu2024buildingmultilinguallanguagemodel} (EN: 10.56B, ZH: 4.35B tokens) and MedRAG \cite{zhao2025medragenhancingretrievalaugmentedgeneration}  (23.9M PubMed snippets) to retain dental-relevant content.


\noindent\textbf{Construction.} We implement a rule-based filtering pipeline using keyword lists derived from TF-IDF analysis on dental and general medical corpora. Starting from vocabularies built on PubMed, MedRAG, and textbook texts, we intersect them with the glossary to obtain candidate dental terms. Terms that appear disproportionately in general medical texts are removed. The final filtering lists contain 440 English and 235 Chinese keywords.

Texts from all sources are filtered using these keywords. We apply a keyword density threshold of >1\% and require at least two distinct matches per sentence. English is tokenized by spaces; Chinese uses direct string matching.

After filtering, we deduplicate the corpus using MD5 hashes, embed texts with the bge-m3 model, and segment into chunks of up to 512 tokens. The final corpus consists of 1.06M English chunks (319.08M tokens) and 66.3K Chinese chunks (18.27M tokens).

\noindent\textbf{Human Validation.}
We manually reviewed 100 random samples per language to assess filter quality, confirming domain relevance rates of 99\% for English and 96\% for Chinese.
\section{Experiments}
\subsection{Experimental Setup}




We split DentalQA into training and test sets in a 4:1 ratio while preserving each subfield’s proportions, and report all results on the held-out test set. MCQ performance is measured by Accuracy; MAQ by Accuracy, Precision, Recall and F1; and OEQ and DEF by BERTScore F1 \citep{zhang2019bertscore}. We conduct our experiments on multiple popular LLMs. For general LLMs, we select DeepSeek-V3, DeepSeek-R1, GPT-4o, GPT-4o-mini, LLaMA-3.2-3B-Instruct, LLaMA-3.1-8B-Instruct \cite{grattafiori2024llama3herdmodels} and Qwen-2.5-1.5/3B/7B/14B/32B-Instruct \cite{qwen2025qwen25technicalreport}. For medical LLMs, we select BioMistral-7B \cite{labrak2024biomistralcollectionopensourcepretrained}, HuatuoGPT2-7B \cite{chen2024huatuogptiionestagetrainingmedical} and LLaMA-3-8B-UltraMedical \cite{UltraMedical}. We evaluate in a zero-shot setting using task-specific prompt templates (Appendix~\ref{sec:eval_prompts}). Experiments are conducted on eight NVIDIA RTX 3090 GPUs.

\begin{table*}[ht]
    \centering
    \caption{Overall Performance on DentalQA. We use Accuracy (ACC), Precistion (P), Recall (R), F1, and BERTScore F1 (BERTScore) as our metrics. \textbf{Bold} indicates the best result, and \underline{underline} indicates the second best.}
    \label{tab:oralkb_bench}
    \small
    \renewcommand{\arraystretch}{1.1}
\setlength{\tabcolsep}{2pt}
        \begin{tabular}{l|c|cccc|c|c|c|c|c}
            \toprule
            & \multicolumn{7}{c|}{\textbf{DentalQA-ZH}} &
            \multicolumn{3}{c}{\textbf{DentalQA-EN}} \\
            & \textbf{MCQ} & \multicolumn{4}{c|}{\textbf{MAQ}} & \textbf{OEQ} & \textbf{DEF} & \textbf{MCQ} & \textbf{OEQ} & \textbf{DEF} \\
            \textbf{Model} & \textbf{ACC} & \textbf{ACC} & \textbf{P} & \textbf{R} & \textbf{F1} & \textbf{BERTScore} & \textbf{BERTScore} & \textbf{ACC} & \textbf{BERTScore} & \textbf{BERTScore}\\
            \midrule
            \multicolumn{11}{l}{\emph{General LLMs}}\\
            \midrule
            GPT-4o          & 64.86 & 37.30 & \underline{87.75} & 81.74 & 84.63 & 27.23 & \underline{21.60} & \textbf{73.98} & 31.28 & 29.21 \\ 
            GPT-4o-mini     & 51.65 & 29.73 & 81.36 & 87.37 & 84.26 & 26.48 & 19.50 & 60.59 & 34.55 & 29.42 \\
            Deepseek-V3     & 69.28 & \underline{41.35} & 87.22 & 86.23 & \underline{86.72} & 27.79 & 17.78 & \underline{68.28} & 27.65 & 25.73 \\
            Deepseek-R1     & \textbf{76.06} & \textbf{43.51} & \textbf{88.64} & 86.68 & \textbf{87.65} & 26.77 & 15.81 & 60.04 & 20.91 & 18.58 \\
            Llama-3.2-3B    & 38.22 &  7.30 & 72.24 & 65.01 & 68.44 & 19.88 & 15.13 & 48.96 & 28.13 & 26.13 \\
            Llama-3.1-8B    & 40.80 & 10.27 & 77.45 & 67.49 & 72.13 & 16.69 & 4.96 & 55.60 & 25.31 & 21.75 \\
            Qwen2.5-1.5B    & 45.58 & 13.24 & 76.67 & 77.55 & 77.11 & 21.74 & 8.57 & 38.09 & 26.10 & 21.59 \\
            Qwen2.5-3B      & 48.63 & 19.19 & 77.70 & 80.37 & 79.01 & 20.89 & 11.16 & 41.77 & 34.48 & 29.62 \\
            Qwen2.5-7B      & 60.29 & 26.22 & 83.08 & 79.22 & 81.11 & 26.37 & 11.59 & 49.23 & 26.28 & 22.12 \\
            Qwen2.5-14B     & 66.48 & 33.51 & 84.05 & 85.01 & 84.53 & 25.47 & 12.69 & 50.49 & 26.68 & 21.93 \\
            Qwen2.5-32B     & \underline{70.86} & 39.46 & 85.50 & 86.15 & 85.82 & 26.02 & 11.65 & 58.34 & 26.59 & 22.78 \\
            \midrule
            \multicolumn{11}{l}{\emph{Medical LLMs}}\\
            \midrule
            BioMistral-7B   & 25.44 & 5.68 & 76.33 & 47.43 & 58.51 & 14.48 & 14.06 & 34.96 & 34.50 & 29.55 \\
            HuatuoGPT2-7B   & 22.51 & 6.22 & 74.50 & 67.54 & 70.85 & 25.38 & 21.04 & 25.47 & 15.50 & 16.55 \\
            Llama-3-8B-UltraMedical & 30.32 & 11.08 & 72.86 & 81.52 & 76.95 & 18.74 & 9.18 & 46.10 & 26.76 & 24.80 \\
            \midrule
            \multicolumn{11}{l}{\emph{Domain Adaptation on Qwen2.5-3B}}\\
            \midrule
            Qwen2.5-3B      & 48.63 & 19.19 & 77.70 & 80.37 & 79.01 & 20.89 & 11.16 & 41.77 & 34.48 & 29.62 \\
            \emph{w}. SFT          & 54.58 & 25.60 & 75.57 & \underline{93.24} & 83.48 & 22.42 & 15.29 & 47.90 & \textbf{37.74} & \textbf{30.79} \\
            \emph{w}. RAG          & 54.45 & 21.35 & 74.88 & 91.17 & 82.22 & \textbf{30.18} & \textbf{22.13} & 48.74 & 36.47 & \underline{30.04} \\
            \emph{w}. SFT+RAG      & 60.06 & 29.07 & 77.30 & \textbf{93.46} & 84.62 & \underline{30.06} & 20.85 & 52.15 & \underline{37.68} & 29.65 \\
            \bottomrule
        \end{tabular}
\end{table*}
\subsection{Domain Adaptation on Qwen2.5-3B}
To enhance dentistry‐specific knowledge and capabilities, we adopt three adaptation strategies based on Qwen-2.5-3B-Instruct. \textbf{(a) Supervised Fine-Tuning (SFT):} Full‐model fine‐tuning on the DentalQA training split for four epochs with a learning rate of 1e-6 and batch size 16 using bfloat16 precision. \textbf{(b) Retrieval-Augmented Generation (RAG):} At inference, retrieve the top-5 most relevant passages from DentalCorpus via FAISS with bge-m3 embeddings and prepend them to the prompt (Appendix~\ref{sec:eval_prompts}). \textbf{(c) SFT + RAG:} Combine the above supervised fine-tuning with retrieval augmentation during inference.

\subsection{Results}
The main results are presented in Table~\ref{tab:oralkb_bench}, where we report the performance of 14 LLMs and our domain adaptation results.

\noindent\textbf{Overall Trends.}  
Performance varies markedly by language. On DentalBench-ZH, DeepSeek-R1 achieves state‐of‐the‐art accuracy on both MCQ and MAQ, with DeepSeek-V3 and Qwen2.5-32B close behind. Conversely, on DentalBench-EN, GPT-4o leads across these tasks. In both languages, however, open-ended tasks (OEQ and DEF) trail far behind MCQ and MAQ, underscoring enduring challenges in domain-specific generative reasoning and terminology.

\noindent\textbf{General Models vs. Medical Models.}  
Although medical LLMs perform relatively well on OEQ and DEF, they fall markedly short of general‐purpose models on MCQ and MAQ. For example, Llama-3.1-8B consistently outperforms its medical counterpart across all multiple-choice tasks, suggesting that medical tuning may insufficiently capture dentistry‐specific factual knowledge.

\noindent\textbf{Impact of Model Scale.}  
In the Qwen-2.5 family, scaling improves MCQ and MAQ notably but yields limited gains on OEQ and DEF, suggesting factual recall benefits more from model size than generative reasoning does.

\noindent\textbf{Domain Adaptation.}  
Both SFT and RAG improve MCQ and MAQ, but RAG shows a larger impact on open-ended tasks (e.g., OEQ-ZH BERTScore: +9.29 vs. +1.53). Combining both yields additive gains—especially on MCQ and MAQ (+11.43 and +9.88). For OEQ/DEF, SFT+RAG offers clear benefit over SFT alone in Chinese, while in English the effect is less consistent, indicating language sensitivity in retrieval effectiveness.

\section{Conclusion}
We introduce DentalBench, a comprehensive bilingual benchmark designed for evaluating and enhancing LLMs in the dental domain. It includes 2 main components: DentalQA, the first bilingual high-quality QA dataset for dentistry, and DentalCorpus, a large-scale domain-specific English-Chinese corpus for domain adaptation, such as SFT and RAG. Our experiments across 14 LLMs, covering proprietary, open-source and medical-specific models, reveal significant performance gaps based on task types, language, and model categories. Additionally, through extensive experiments, we demonstrate that domain adaptation using DentalCorpus can significantly improve performance. In general, DentalBench can be served as a valuable resource for evaluating knowledge-grounded language models in dentistry, improving language understanding in oral healthcare, and encouraging more related research.


\section*{Limitations}
Our work has several limitations. First, the dataset exhibits asymmetry between Chinese and English sources. While both languages are supported throughout DentalQA and DentalCorpus, the distribution, source diversity, and depth of coverage are not fully aligned—potentially contributing to observed cross-lingual performance gaps. Second, the MAQ format is currently only available in Chinese, limiting comprehensive evaluation of multi-answer reasoning capabilities in English. In future work, we aim to construct balanced bilingual resources and expand task coverage across languages.

\bibliography{main}

\begin{thebibliography}{32}
\providecommand{\natexlab}[1]{#1}

\bibitem[{{Ben Abacha} and Demner{-}Fushman(2019)}]{BenAbacha-BMC-2019}
Asma {Ben Abacha} and Dina Demner{-}Fushman. 2019.
\newblock \href {https://bmcbioinformatics.biomedcentral.com/articles/10.1186/s12859-019-3119-4} {A question-entailment approach to question answering}.
\newblock \emph{{BMC} Bioinform.}, 20(1):511:1--511:23.

\bibitem[{Chen et~al.(2024)Chen, Wang, Ji, Gao, Jiang, Chen, Zhang, Song, Xie, Kong, Li, Wan, Li, and Wang}]{chen2024huatuogptiionestagetrainingmedical}
Junying Chen, Xidong Wang, Ke~Ji, Anningzhe Gao, Feng Jiang, Shunian Chen, Hongbo Zhang, Dingjie Song, Wenya Xie, Chuyi Kong, Jianquan Li, Xiang Wan, Haizhou Li, and Benyou Wang. 2024.
\newblock \href {https://arxiv.org/abs/2311.09774} {Huatuogpt-ii, one-stage training for medical adaption of llms}.
\newblock \emph{Preprint}, arXiv:2311.09774.

\bibitem[{DeepSeek-AI et~al.(2025)DeepSeek-AI, Guo, Yang, Zhang, Song, Zhang, Xu, Zhu, Ma, Wang, Bi, Zhang, Yu, Wu, Wu, Gou, Shao, Li, Gao, Liu, Xue, Wang, Wu, Feng, Lu, Zhao, Deng, Zhang, Ruan, Dai, Chen, Ji, Li, Lin, Dai, Luo, Hao, Chen, Li, Zhang, Bao, Xu, Wang, Ding, Xin, Gao, Qu, Li, Guo, Li, Wang, Chen, Yuan, Qiu, Li, Cai, Ni, Liang, Chen, Dong, Hu, Gao, Guan, Huang, Yu, Wang, Zhang, Zhao, Wang, Zhang, Xu, Xia, Zhang, Zhang, Tang, Li, Wang, Li, Tian, Huang, Zhang, Wang, Chen, Du, Ge, Zhang, Pan, Wang, Chen, Jin, Chen, Lu, Zhou, Chen, Ye, Wang, Yu, Zhou, Pan, Li, Zhou, Wu, Ye, Yun, Pei, Sun, Wang, Zeng, Zhao, Liu, Liang, Gao, Yu, Zhang, Xiao, An, Liu, Wang, Chen, Nie, Cheng, Liu, Xie, Liu, Yang, Li, Su, Lin, Li, Jin, Shen, Chen, Sun, Wang, Song, Zhou, Wang, Shan, Li, Wang, Wei, Zhang, Xu, Li, Zhao, Sun, Wang, Yu, Zhang, Shi, Xiong, He, Piao, Wang, Tan, Ma, Liu, Guo, Ou, Wang, Gong, Zou, He, Xiong, Luo, You, Liu, Zhou, Zhu, Xu, Huang, Li, Zheng, Zhu, Ma, Tang, Zha, Yan, Ren, Ren, Sha, Fu, Xu, Xie, Zhang,
  Hao, Ma, Yan, Wu, Gu, Zhu, Liu, Li, Xie, Song, Pan, Huang, Xu, Zhang, and Zhang}]{deepseekai2025deepseekr1incentivizingreasoningcapability}
DeepSeek-AI, Daya Guo, Dejian Yang, Haowei Zhang, Junxiao Song, Ruoyu Zhang, Runxin Xu, Qihao Zhu, Shirong Ma, Peiyi Wang, Xiao Bi, Xiaokang Zhang, Xingkai Yu, Yu~Wu, Z.~F. Wu, Zhibin Gou, Zhihong Shao, Zhuoshu Li, Ziyi Gao, and 181 others. 2025.
\newblock \href {https://arxiv.org/abs/2501.12948} {Deepseek-r1: Incentivizing reasoning capability in llms via reinforcement learning}.
\newblock \emph{Preprint}, arXiv:2501.12948.

\bibitem[{Grattafiori et~al.(2024)Grattafiori, Dubey, Jauhri, Pandey, Kadian, Al-Dahle, Letman, Mathur, Schelten, Vaughan, Yang, Fan, Goyal, Hartshorn, Yang, Mitra, Sravankumar, Korenev, Hinsvark, Rao, Zhang, Rodriguez, Gregerson, Spataru, Roziere, Biron, Tang, Chern, Caucheteux, Nayak, Bi, Marra, McConnell, Keller, Touret, Wu, Wong, Ferrer, Nikolaidis, Allonsius, Song, Pintz, Livshits, Wyatt, Esiobu, Choudhary, Mahajan, Garcia-Olano, Perino, Hupkes, Lakomkin, AlBadawy, Lobanova, Dinan, Smith, Radenovic, Guzmán, Zhang, Synnaeve, Lee, Anderson, Thattai, Nail, Mialon, Pang, Cucurell, Nguyen, Korevaar, Xu, Touvron, Zarov, Ibarra, Kloumann, Misra, Evtimov, Zhang, Copet, Lee, Geffert, Vranes, Park, Mahadeokar, Shah, van~der Linde, Billock, Hong, Lee, Fu, Chi, Huang, Liu, Wang, Yu, Bitton, Spisak, Park, Rocca, Johnstun, Saxe, Jia, Alwala, Prasad, Upasani, Plawiak, Li, Heafield, Stone, El-Arini, Iyer, Malik, Chiu, Bhalla, Lakhotia, Rantala-Yeary, van~der Maaten, Chen, Tan, Jenkins, Martin, Madaan, Malo, Blecher,
  Landzaat, de~Oliveira, Muzzi, Pasupuleti, Singh, Paluri, Kardas, Tsimpoukelli, Oldham, Rita, Pavlova, Kambadur, Lewis, Si, Singh, Hassan, Goyal, Torabi, Bashlykov, Bogoychev, Chatterji, Zhang, Duchenne, Çelebi, Alrassy, Zhang, Li, Vasic, Weng, Bhargava, Dubal, Krishnan, Koura, Xu, He, Dong, Srinivasan, Ganapathy, Calderer, Cabral, Stojnic, Raileanu, Maheswari, Girdhar, Patel, Sauvestre, Polidoro, Sumbaly, Taylor, Silva, Hou, Wang, Hosseini, Chennabasappa, Singh, Bell, Kim, Edunov, Nie, Narang, Raparthy, Shen, Wan, Bhosale, Zhang, Vandenhende, Batra, Whitman, Sootla, Collot, Gururangan, Borodinsky, Herman, Fowler, Sheasha, Georgiou, Scialom, Speckbacher, Mihaylov, Xiao, Karn, Goswami, Gupta, Ramanathan, Kerkez, Gonguet, Do, Vogeti, Albiero, Petrovic, Chu, Xiong, Fu, Meers, Martinet, Wang, Wang, Tan, Xia, Xie, Jia, Wang, Goldschlag, Gaur, Babaei, Wen, Song, Zhang, Li, Mao, Coudert, Yan, Chen, Papakipos, Singh, Srivastava, Jain, Kelsey, Shajnfeld, Gangidi, Victoria, Goldstand, Menon, Sharma, Boesenberg,
  Baevski, Feinstein, Kallet, Sangani, Teo, Yunus, Lupu, Alvarado, Caples, Gu, Ho, Poulton, Ryan, Ramchandani, Dong, Franco, Goyal, Saraf, Chowdhury, Gabriel, Bharambe, Eisenman, Yazdan, James, Maurer, Leonhardi, Huang, Loyd, Paola, Paranjape, Liu, Wu, Ni, Hancock, Wasti, Spence, Stojkovic, Gamido, Montalvo, Parker, Burton, Mejia, Liu, Wang, Kim, Zhou, Hu, Chu, Cai, Tindal, Feichtenhofer, Gao, Civin, Beaty, Kreymer, Li, Adkins, Xu, Testuggine, David, Parikh, Liskovich, Foss, Wang, Le, Holland, Dowling, Jamil, Montgomery, Presani, Hahn, Wood, Le, Brinkman, Arcaute, Dunbar, Smothers, Sun, Kreuk, Tian, Kokkinos, Ozgenel, Caggioni, Kanayet, Seide, Florez, Schwarz, Badeer, Swee, Halpern, Herman, Sizov, Guangyi, Zhang, Lakshminarayanan, Inan, Shojanazeri, Zou, Wang, Zha, Habeeb, Rudolph, Suk, Aspegren, Goldman, Zhan, Damlaj, Molybog, Tufanov, Leontiadis, Veliche, Gat, Weissman, Geboski, Kohli, Lam, Asher, Gaya, Marcus, Tang, Chan, Zhen, Reizenstein, Teboul, Zhong, Jin, Yang, Cummings, Carvill, Shepard, McPhie,
  Torres, Ginsburg, Wang, Wu, U, Saxena, Khandelwal, Zand, Matosich, Veeraraghavan, Michelena, Li, Jagadeesh, Huang, Chawla, Huang, Chen, Garg, A, Silva, Bell, Zhang, Guo, Yu, Moshkovich, Wehrstedt, Khabsa, Avalani, Bhatt, Mankus, Hasson, Lennie, Reso, Groshev, Naumov, Lathi, Keneally, Liu, Seltzer, Valko, Restrepo, Patel, Vyatskov, Samvelyan, Clark, Macey, Wang, Hermoso, Metanat, Rastegari, Bansal, Santhanam, Parks, White, Bawa, Singhal, Egebo, Usunier, Mehta, Laptev, Dong, Cheng, Chernoguz, Hart, Salpekar, Kalinli, Kent, Parekh, Saab, Balaji, Rittner, Bontrager, Roux, Dollar, Zvyagina, Ratanchandani, Yuvraj, Liang, Alao, Rodriguez, Ayub, Murthy, Nayani, Mitra, Parthasarathy, Li, Hogan, Battey, Wang, Howes, Rinott, Mehta, Siby, Bondu, Datta, Chugh, Hunt, Dhillon, Sidorov, Pan, Mahajan, Verma, Yamamoto, Ramaswamy, Lindsay, Lindsay, Feng, Lin, Zha, Patil, Shankar, Zhang, Zhang, Wang, Agarwal, Sajuyigbe, Chintala, Max, Chen, Kehoe, Satterfield, Govindaprasad, Gupta, Deng, Cho, Virk, Subramanian, Choudhury,
  Goldman, Remez, Glaser, Best, Koehler, Robinson, Li, Zhang, Matthews, Chou, Shaked, Vontimitta, Ajayi, Montanez, Mohan, Kumar, Mangla, Ionescu, Poenaru, Mihailescu, Ivanov, Li, Wang, Jiang, Bouaziz, Constable, Tang, Wu, Wang, Wu, Gao, Kleinman, Chen, Hu, Jia, Qi, Li, Zhang, Zhang, Adi, Nam, Yu, Wang, Zhao, Hao, Qian, Li, He, Rait, DeVito, Rosnbrick, Wen, Yang, Zhao, and Ma}]{grattafiori2024llama3herdmodels}
Aaron Grattafiori, Abhimanyu Dubey, Abhinav Jauhri, Abhinav Pandey, Abhishek Kadian, Ahmad Al-Dahle, Aiesha Letman, Akhil Mathur, Alan Schelten, Alex Vaughan, Amy Yang, Angela Fan, Anirudh Goyal, Anthony Hartshorn, Aobo Yang, Archi Mitra, Archie Sravankumar, Artem Korenev, Arthur Hinsvark, and 542 others. 2024.
\newblock \href {https://arxiv.org/abs/2407.21783} {The llama 3 herd of models}.
\newblock \emph{Preprint}, arXiv:2407.21783.

\bibitem[{Hendrycks et~al.(2021)Hendrycks, Burns, Basart, Zou, Mazeika, Song, and Steinhardt}]{hendrycks2021measuringmassivemultitasklanguage}
Dan Hendrycks, Collin Burns, Steven Basart, Andy Zou, Mantas Mazeika, Dawn Song, and Jacob Steinhardt. 2021.
\newblock \href {https://arxiv.org/abs/2009.03300} {Measuring massive multitask language understanding}.
\newblock \emph{Preprint}, arXiv:2009.03300.

\bibitem[{Jaech et~al.(2024)Jaech, Kalai, Lerer, Richardson, El-Kishky, Low, Helyar, Madry, Beutel, Carney et~al.}]{openaio1}
Aaron Jaech, Adam Kalai, Adam Lerer, Adam Richardson, Ahmed El-Kishky, Aiden Low, Alec Helyar, Aleksander Madry, Alex Beutel, Alex Carney, and 1 others. 2024.
\newblock Openai o1 system card.
\newblock \emph{arXiv preprint arXiv:2412.16720}.

\bibitem[{Jin et~al.(2020)Jin, Pan, Oufattole, Weng, Fang, and Szolovits}]{jin2020diseasedoespatienthave}
Di~Jin, Eileen Pan, Nassim Oufattole, Wei-Hung Weng, Hanyi Fang, and Peter Szolovits. 2020.
\newblock \href {https://arxiv.org/abs/2009.13081} {What disease does this patient have? a large-scale open domain question answering dataset from medical exams}.
\newblock \emph{Preprint}, arXiv:2009.13081.

\bibitem[{Jin et~al.(2019)Jin, Dhingra, Liu, Cohen, and Lu}]{jin2019pubmedqadatasetbiomedicalresearch}
Qiao Jin, Bhuwan Dhingra, Zhengping Liu, William~W. Cohen, and Xinghua Lu. 2019.
\newblock \href {https://arxiv.org/abs/1909.06146} {Pubmedqa: A dataset for biomedical research question answering}.
\newblock \emph{Preprint}, arXiv:1909.06146.

\bibitem[{Labrak et~al.(2024)Labrak, Bazoge, Morin, Gourraud, Rouvier, and Dufour}]{labrak2024biomistralcollectionopensourcepretrained}
Yanis Labrak, Adrien Bazoge, Emmanuel Morin, Pierre-Antoine Gourraud, Mickael Rouvier, and Richard Dufour. 2024.
\newblock \href {https://arxiv.org/abs/2402.10373} {Biomistral: A collection of open-source pretrained large language models for medical domains}.
\newblock \emph{Preprint}, arXiv:2402.10373.

\bibitem[{Li et~al.(2023)Li, Wong, Zhang, Usuyama, Liu, Yang, Naumann, Poon, and Gao}]{llavamed}
Chunyuan Li, Cliff Wong, Sheng Zhang, Naoto Usuyama, Haotian Liu, Jianwei Yang, Tristan Naumann, Hoifung Poon, and Jianfeng Gao. 2023.
\newblock Llava-med: Training a large language-and-vision assistant for biomedicine in one day.
\newblock \emph{Advances in Neural Information Processing Systems}, 36:28541--28564.

\bibitem[{Liu et~al.(2024)Liu, Feng, Xue, Wang, Wu, Lu, Zhao, Deng, Zhang, Ruan et~al.}]{liu2024deepseek}
Aixin Liu, Bei Feng, Bing Xue, Bingxuan Wang, Bochao Wu, Chengda Lu, Chenggang Zhao, Chengqi Deng, Chenyu Zhang, Chong Ruan, and 1 others. 2024.
\newblock Deepseek-v3 technical report.
\newblock \emph{arXiv preprint arXiv:2412.19437}.

\bibitem[{Liu et~al.(2023)Liu, Hao, Lin, Pan, Yang, Feng, Wang, Li, Jin, Zhao et~al.}]{liu2023deep}
Jiaxiang Liu, Jin Hao, Hangzheng Lin, Wei Pan, Jianfei Yang, Yang Feng, Gaoang Wang, Jin Li, Zuolin Jin, Zhihe Zhao, and 1 others. 2023.
\newblock Deep learning-enabled 3d multimodal fusion of cone-beam ct and intraoral mesh scans for clinically applicable tooth-bone reconstruction.
\newblock \emph{Patterns}, 4(9).

\bibitem[{OpenAI et~al.(2024)OpenAI, Achiam, Adler, Agarwal, Ahmad, Akkaya, Aleman, Almeida, Altenschmidt, Altman, Anadkat, Avila, Babuschkin, Balaji, Balcom, Baltescu, Bao, Bavarian, Belgum, Bello, Berdine, Bernadett-Shapiro, Berner, Bogdonoff, Boiko, Boyd, Brakman, Brockman, Brooks, Brundage, Button, Cai, Campbell, Cann, Carey, Carlson, Carmichael, Chan, Chang, Chantzis, Chen, Chen, Chen, Chen, Chen, Chess, Cho, Chu, Chung, Cummings, Currier, Dai, Decareaux, Degry, Deutsch, Deville, Dhar, Dohan, Dowling, Dunning, Ecoffet, Eleti, Eloundou, Farhi, Fedus, Felix, Fishman, Forte, Fulford, Gao, Georges, Gibson, Goel, Gogineni, Goh, Gontijo-Lopes, Gordon, Grafstein, Gray, Greene, Gross, Gu, Guo, Hallacy, Han, Harris, He, Heaton, Heidecke, Hesse, Hickey, Hickey, Hoeschele, Houghton, Hsu, Hu, Hu, Huizinga, Jain, Jain, Jang, Jiang, Jiang, Jin, Jin, Jomoto, Jonn, Jun, Kaftan, Łukasz Kaiser, Kamali, Kanitscheider, Keskar, Khan, Kilpatrick, Kim, Kim, Kim, Kirchner, Kiros, Knight, Kokotajlo, Łukasz Kondraciuk,
  Kondrich, Konstantinidis, Kosic, Krueger, Kuo, Lampe, Lan, Lee, Leike, Leung, Levy, Li, Lim, Lin, Lin, Litwin, Lopez, Lowe, Lue, Makanju, Malfacini, Manning, Markov, Markovski, Martin, Mayer, Mayne, McGrew, McKinney, McLeavey, McMillan, McNeil, Medina, Mehta, Menick, Metz, Mishchenko, Mishkin, Monaco, Morikawa, Mossing, Mu, Murati, Murk, Mély, Nair, Nakano, Nayak, Neelakantan, Ngo, Noh, Ouyang, O'Keefe, Pachocki, Paino, Palermo, Pantuliano, Parascandolo, Parish, Parparita, Passos, Pavlov, Peng, Perelman, de~Avila Belbute~Peres, Petrov, de~Oliveira~Pinto, Michael, Pokorny, Pokrass, Pong, Powell, Power, Power, Proehl, Puri, Radford, Rae, Ramesh, Raymond, Real, Rimbach, Ross, Rotsted, Roussez, Ryder, Saltarelli, Sanders, Santurkar, Sastry, Schmidt, Schnurr, Schulman, Selsam, Sheppard, Sherbakov, Shieh, Shoker, Shyam, Sidor, Sigler, Simens, Sitkin, Slama, Sohl, Sokolowsky, Song, Staudacher, Such, Summers, Sutskever, Tang, Tezak, Thompson, Tillet, Tootoonchian, Tseng, Tuggle, Turley, Tworek, Uribe, Vallone,
  Vijayvergiya, Voss, Wainwright, Wang, Wang, Wang, Ward, Wei, Weinmann, Welihinda, Welinder, Weng, Weng, Wiethoff, Willner, Winter, Wolrich, Wong, Workman, Wu, Wu, Wu, Xiao, Xu, Yoo, Yu, Yuan, Zaremba, Zellers, Zhang, Zhang, Zhao, Zheng, Zhuang, Zhuk, and Zoph}]{openai2024gpt4technicalreport}
OpenAI, Josh Achiam, Steven Adler, Sandhini Agarwal, Lama Ahmad, Ilge Akkaya, Florencia~Leoni Aleman, Diogo Almeida, Janko Altenschmidt, Sam Altman, Shyamal Anadkat, Red Avila, Igor Babuschkin, Suchir Balaji, Valerie Balcom, Paul Baltescu, Haiming Bao, Mohammad Bavarian, Jeff Belgum, and 262 others. 2024.
\newblock \href {https://arxiv.org/abs/2303.08774} {Gpt-4 technical report}.
\newblock \emph{Preprint}, arXiv:2303.08774.

\bibitem[{Pal et~al.(2022)Pal, Umapathi, and Sankarasubbu}]{pal2022medmcqalargescalemultisubject}
Ankit Pal, Logesh~Kumar Umapathi, and Malaikannan Sankarasubbu. 2022.
\newblock \href {https://arxiv.org/abs/2203.14371} {Medmcqa : A large-scale multi-subject multi-choice dataset for medical domain question answering}.
\newblock \emph{Preprint}, arXiv:2203.14371.

\bibitem[{Qiu et~al.(2024)Qiu, Wu, Zhang, Lin, Wang, Zhang, Wang, and Xie}]{qiu2024buildingmultilinguallanguagemodel}
Pengcheng Qiu, Chaoyi Wu, Xiaoman Zhang, Weixiong Lin, Haicheng Wang, Ya~Zhang, Yanfeng Wang, and Weidi Xie. 2024.
\newblock \href {https://arxiv.org/abs/2402.13963} {Towards building multilingual language model for medicine}.
\newblock \emph{Preprint}, arXiv:2402.13963.

\bibitem[{Qwen et~al.(2025)Qwen, :, Yang, Yang, Zhang, Hui, Zheng, Yu, Li, Liu, Huang, Wei, Lin, Yang, Tu, Zhang, Yang, Yang, Zhou, Lin, Dang, Lu, Bao, Yang, Yu, Li, Xue, Zhang, Zhu, Men, Lin, Li, Tang, Xia, Ren, Ren, Fan, Su, Zhang, Wan, Liu, Cui, Zhang, and Qiu}]{qwen2025qwen25technicalreport}
Qwen, :, An~Yang, Baosong Yang, Beichen Zhang, Binyuan Hui, Bo~Zheng, Bowen Yu, Chengyuan Li, Dayiheng Liu, Fei Huang, Haoran Wei, Huan Lin, Jian Yang, Jianhong Tu, Jianwei Zhang, Jianxin Yang, Jiaxi Yang, Jingren Zhou, and 25 others. 2025.
\newblock \href {https://arxiv.org/abs/2412.15115} {Qwen2.5 technical report}.
\newblock \emph{Preprint}, arXiv:2412.15115.

\bibitem[{Regin(2017)}]{regin2017medical}
Lasse Regin. 2017.
\newblock Medical question answer data.
\newblock \url{https://github.com/LasseRegin/medical-question-answer-data}.
\newblock Accessed: May 15, 2023.

\bibitem[{Shi et~al.(2024)Shi, Meng, Chen, Feng, Zhao, Hao, Fang, Liu, and Zheng}]{shi2024leta}
Zefeng Shi, Zijie Meng, Ruizhe Chen, Yang Feng, Zeyu Zhao, Jin Hao, Bing Fang, Zuozhu Liu, and Youyi Zheng. 2024.
\newblock Leta: Tooth alignment prediction based on dual-branch latent encoding.
\newblock \emph{IEEE Transactions on Visualization and Computer Graphics}.

\bibitem[{Team et~al.(2025)Team, Kamath, Ferret, Pathak, Vieillard, Merhej, Perrin, Matejovicova, Ram{\'e}, Rivi{\`e}re et~al.}]{team2025gemma}
Gemma Team, Aishwarya Kamath, Johan Ferret, Shreya Pathak, Nino Vieillard, Ramona Merhej, Sarah Perrin, Tatiana Matejovicova, Alexandre Ram{\'e}, Morgane Rivi{\`e}re, and 1 others. 2025.
\newblock Gemma 3 technical report.
\newblock \emph{arXiv preprint arXiv:2503.19786}.

\bibitem[{{U.S. National Library of Medicine}(2025{\natexlab{a}})}]{pubmed}
{U.S. National Library of Medicine}. 2025{\natexlab{a}}.
\newblock Pubmed: Medline retrieval system.
\newblock \url{https://pubmed.ncbi.nlm.nih.gov/}.
\newblock Accessed: 2025-05-20.

\bibitem[{{U.S. National Library of Medicine}(2025{\natexlab{b}})}]{umls_nlm}
{U.S. National Library of Medicine}. 2025{\natexlab{b}}.
\newblock Unified medical language system (umls).
\newblock \url{https://www.nlm.nih.gov/research/umls}.
\newblock Accessed: 2025-05-20.

\bibitem[{Wei et~al.(2020)Wei, Cui, Liu, Chen, Chen, Li, and Wang}]{wei2020tanet}
Guodong Wei, Zhiming Cui, Yumeng Liu, Nenglun Chen, Runnan Chen, Guiqing Li, and Wenping Wang. 2020.
\newblock Tanet: towards fully automatic tooth arrangement.
\newblock In \emph{Computer Vision--ECCV 2020: 16th European Conference, Glasgow, UK, August 23--28, 2020, Proceedings, Part XV 16}, pages 481--497. Springer.

\bibitem[{Wu et~al.(2024)Wu, Lin, Zhang, Zhang, Xie, and Wang}]{pmcllama}
Chaoyi Wu, Weixiong Lin, Xiaoman Zhang, Ya~Zhang, Weidi Xie, and Yanfeng Wang. 2024.
\newblock Pmc-llama: toward building open-source language models for medicine.
\newblock \emph{Journal of the American Medical Informatics Association}, 31(9):1833--1843.

\bibitem[{Wu et~al.(2025)Wu, Qiu, Liu, Gu, Li, Zhang, Wang, and Xie}]{wu2025towards}
Chaoyi Wu, Pengcheng Qiu, Jinxin Liu, Hongfei Gu, Na~Li, Ya~Zhang, Yanfeng Wang, and Weidi Xie. 2025.
\newblock Towards evaluating and building versatile large language models for medicine.
\newblock \emph{npj Digital Medicine}, 8(1):58.

\bibitem[{Xiong et~al.(2023)Xiong, Li, Tan, Feng, Zhou, Hao, Ying, Wu, and Liu}]{xiong2023tsegformer}
Huimin Xiong, Kunle Li, Kaiyuan Tan, Yang Feng, Joey~Tianyi Zhou, Jin Hao, Haochao Ying, Jian Wu, and Zuozhu Liu. 2023.
\newblock Tsegformer: 3d tooth segmentation in intraoral scans with geometry guided transformer.
\newblock In \emph{International Conference on Medical Image Computing and Computer-Assisted Intervention}, pages 421--432. Springer.

\bibitem[{Yang et~al.(2025)Yang, Li, Yang, Zhang, Hui, Zheng, Yu, Gao, Huang, Lv et~al.}]{yang2025qwen3}
An~Yang, Anfeng Li, Baosong Yang, Beichen Zhang, Binyuan Hui, Bo~Zheng, Bowen Yu, Chang Gao, Chengen Huang, Chenxu Lv, and 1 others. 2025.
\newblock Qwen3 technical report.
\newblock \emph{arXiv preprint arXiv:2505.09388}.

\bibitem[{Yu et~al.(2024)Yu, Yu, Wang, Zou, and Qin}]{yu2024enhancinghealthcarelargelanguage}
Haoran Yu, Chang Yu, Zihan Wang, Dongxian Zou, and Hao Qin. 2024.
\newblock \href {https://arxiv.org/abs/2408.04138} {Enhancing healthcare through large language models: A study on medical question answering}.
\newblock \emph{Preprint}, arXiv:2408.04138.

\bibitem[{Zhang et~al.(2024)Zhang, Ding, Qi, Zeng, Li, Zhu, Chen, and Zhou}]{UltraMedical}
Kaiyan Zhang, Ning Ding, Biqing Qi, Sihang Zeng, Haoxin Li, Xuekai Zhu, Zhang-Ren Chen, and Bowen Zhou. 2024.
\newblock Ultramedical: Building specialized generalists in biomedicine.
\newblock \url{https://github.com/TsinghuaC3I/UltraMedical}.

\bibitem[{Zhang et~al.(2019)Zhang, Kishore, Wu, Weinberger, and Artzi}]{zhang2019bertscore}
Tianyi Zhang, Varsha Kishore, Felix Wu, Kilian~Q Weinberger, and Yoav Artzi. 2019.
\newblock Bertscore: Evaluating text generation with bert.
\newblock \emph{arXiv preprint arXiv:1904.09675}.

\bibitem[{Zhao et~al.(2025)Zhao, Liu, Yang, and Miao}]{zhao2025medragenhancingretrievalaugmentedgeneration}
Xuejiao Zhao, Siyan Liu, Su-Yin Yang, and Chunyan Miao. 2025.
\newblock \href {https://arxiv.org/abs/2502.04413} {Medrag: Enhancing retrieval-augmented generation with knowledge graph-elicited reasoning for healthcare copilot}.
\newblock \emph{Preprint}, arXiv:2502.04413.

\bibitem[{Zhao et~al.(2020)Zhao, Zhou, and Bai}]{zhao2020orthodontics}
Zhihe Zhao, Yanheng Zhou, and Yuxing Bai. 2020.
\newblock \emph{Orthodontics}.
\newblock People's Medical Publishing House, Beijing.
\newblock In Chinese.

\bibitem[{Zhou et~al.(2023)Zhou, Chia, Wagner, Ayhan, Williamson, Struyven, Liu, Xu, Lozano, Woodward-Court et~al.}]{retfound}
Yukun Zhou, Mark~A Chia, Siegfried~K Wagner, Murat~S Ayhan, Dominic~J Williamson, Robbert~R Struyven, Timing Liu, Moucheng Xu, Mateo~G Lozano, Peter Woodward-Court, and 1 others. 2023.
\newblock A foundation model for generalizable disease detection from retinal images.
\newblock \emph{Nature}, 622(7981):156--163.

\end{thebibliography}

\appendix

\section{Dataset Construction Prompts \& Templates}
\label{sec:Prompts}

\subsection{Definition Templates}
\label{sec:def_templates}
\begin{figure*}
    \centering
    \includegraphics[width=1\linewidth]{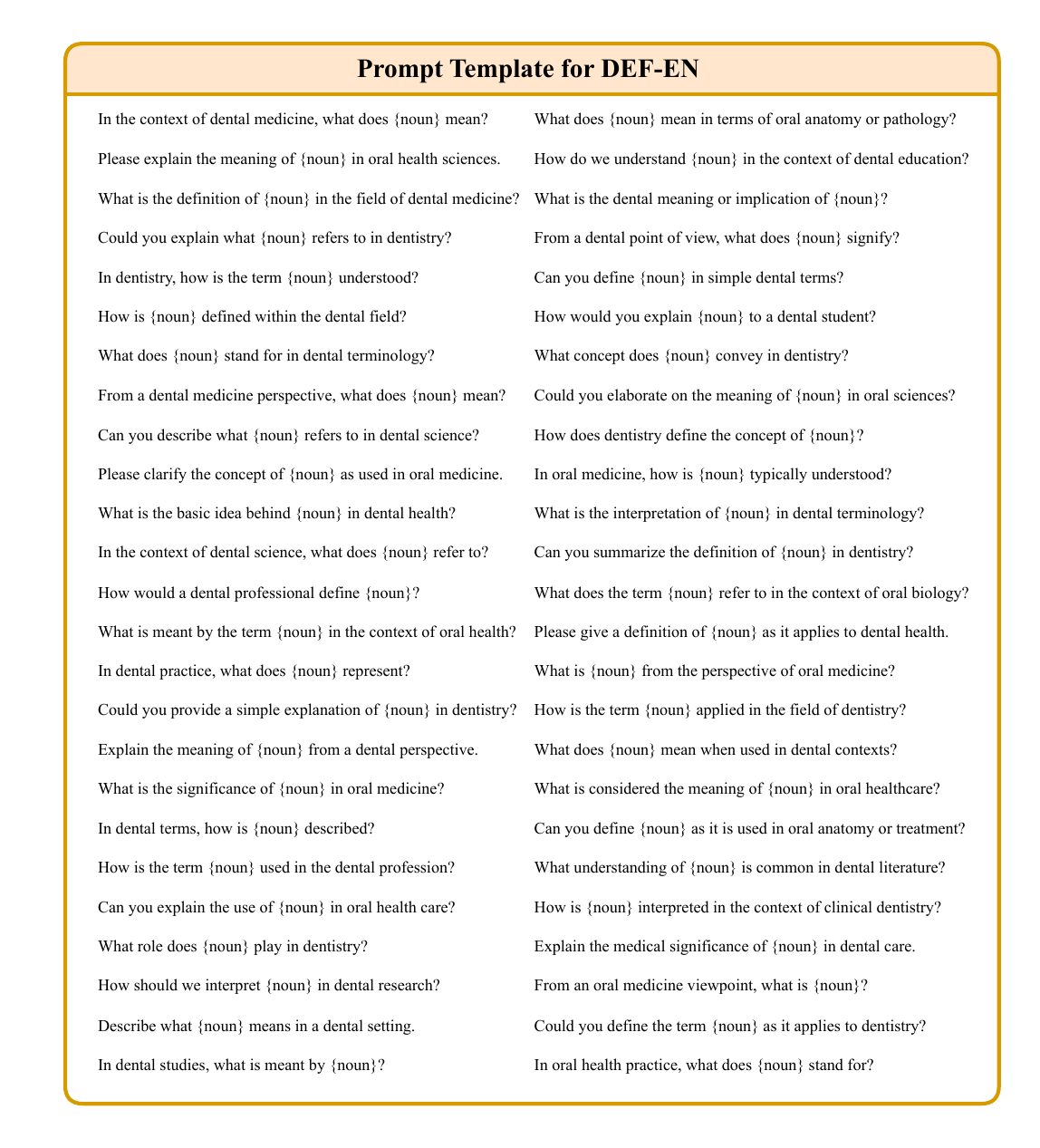}
    \caption{Templates for DEF-EN}
    \label{fig:DEF-EN}
    \vspace{-1em}
\end{figure*}

\begin{figure*}
    \centering
    \includegraphics[width=1\linewidth]{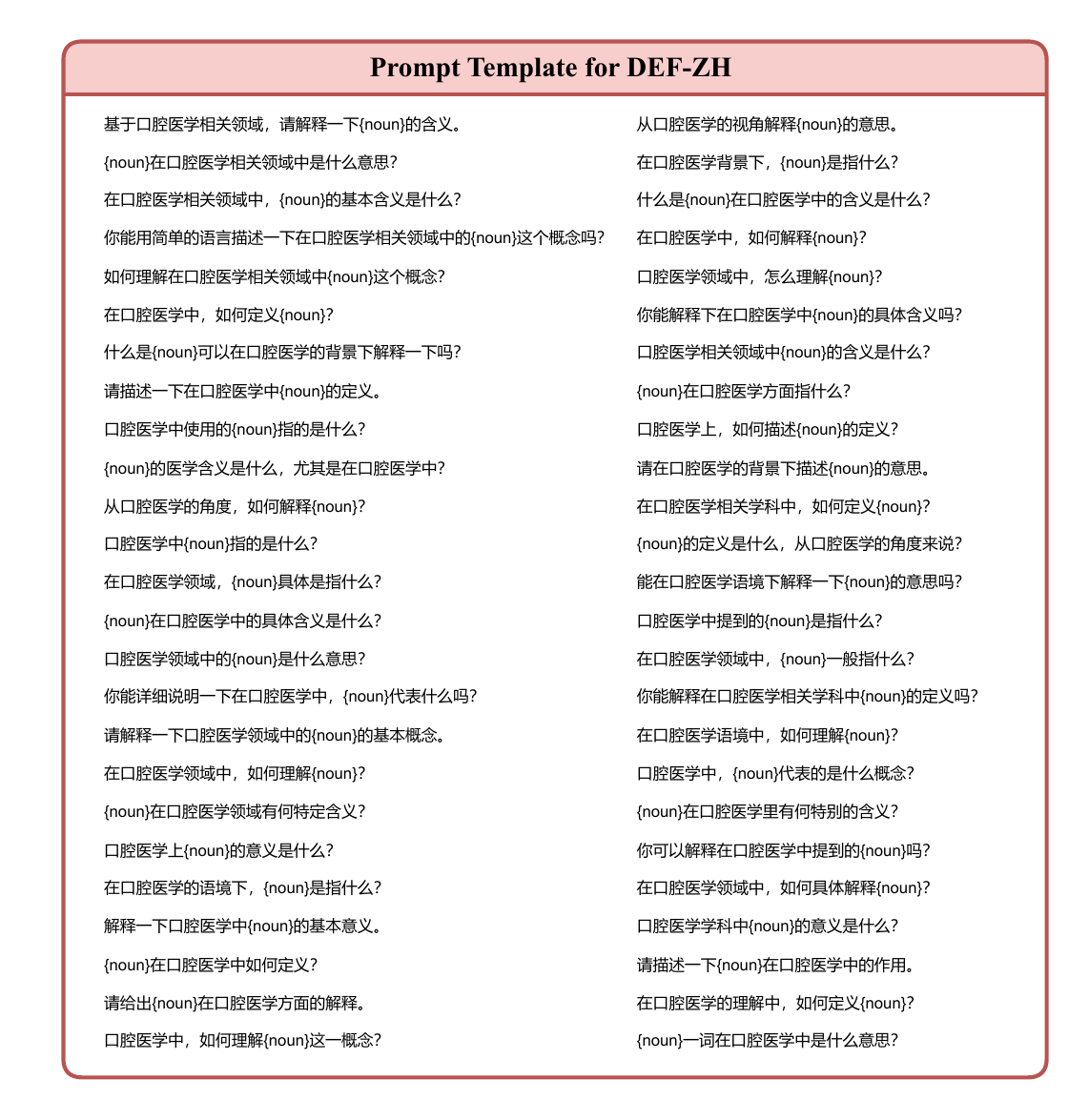}
    \caption{Templates for DEF-ZH}
    \label{fig:DEF-ZH}
    \vspace{-1em}
\end{figure*}

Fig.~\ref{fig:DEF-EN} and Fig.~\ref{fig:DEF-ZH} list the 50 instruction templates used to construct DEF questions from domain terms in English and Chinese.

\subsection{Filtering Classification Prompt}
\label{sec:classification_prompt}
\begin{figure*}
    \centering
    \includegraphics[width=1\linewidth]{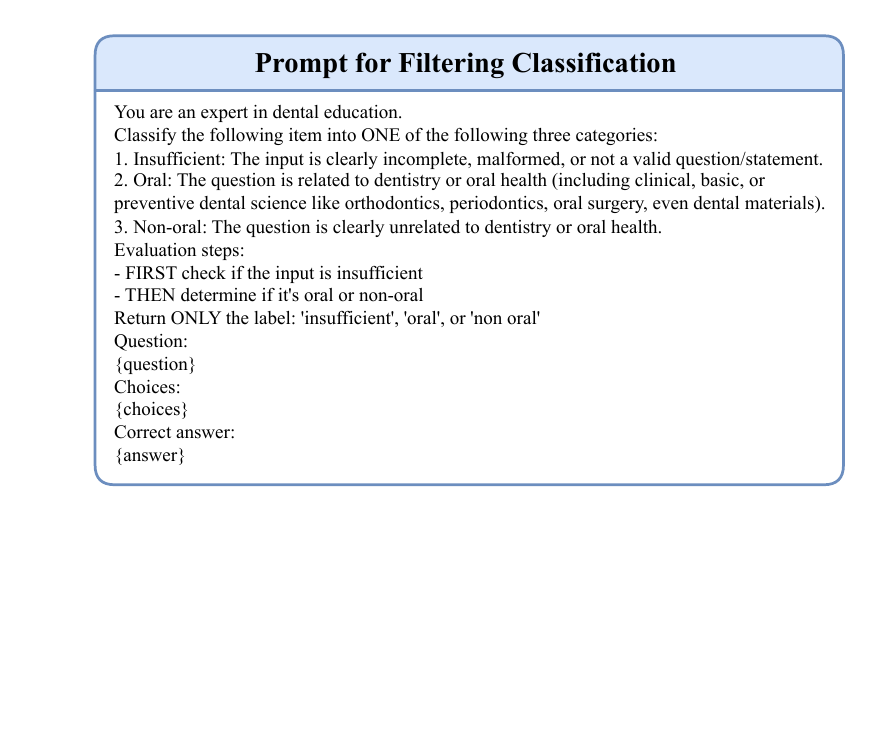}
    \caption{Filtering Classification Prompt}
    \label{fig:Filtering_Classification_Prompt}
\end{figure*}

Fig.~\ref{fig:Filtering_Classification_Prompt} shows the prompt for classifying questions into \textit{oral-related}, \textit{non-oral}, or \textit{insufficient} categories.

\subsection{Evaluation and RAG Prompts}
\label{sec:eval_prompts}
\begin{figure*}
    \centering
    \includegraphics[width=1\linewidth]{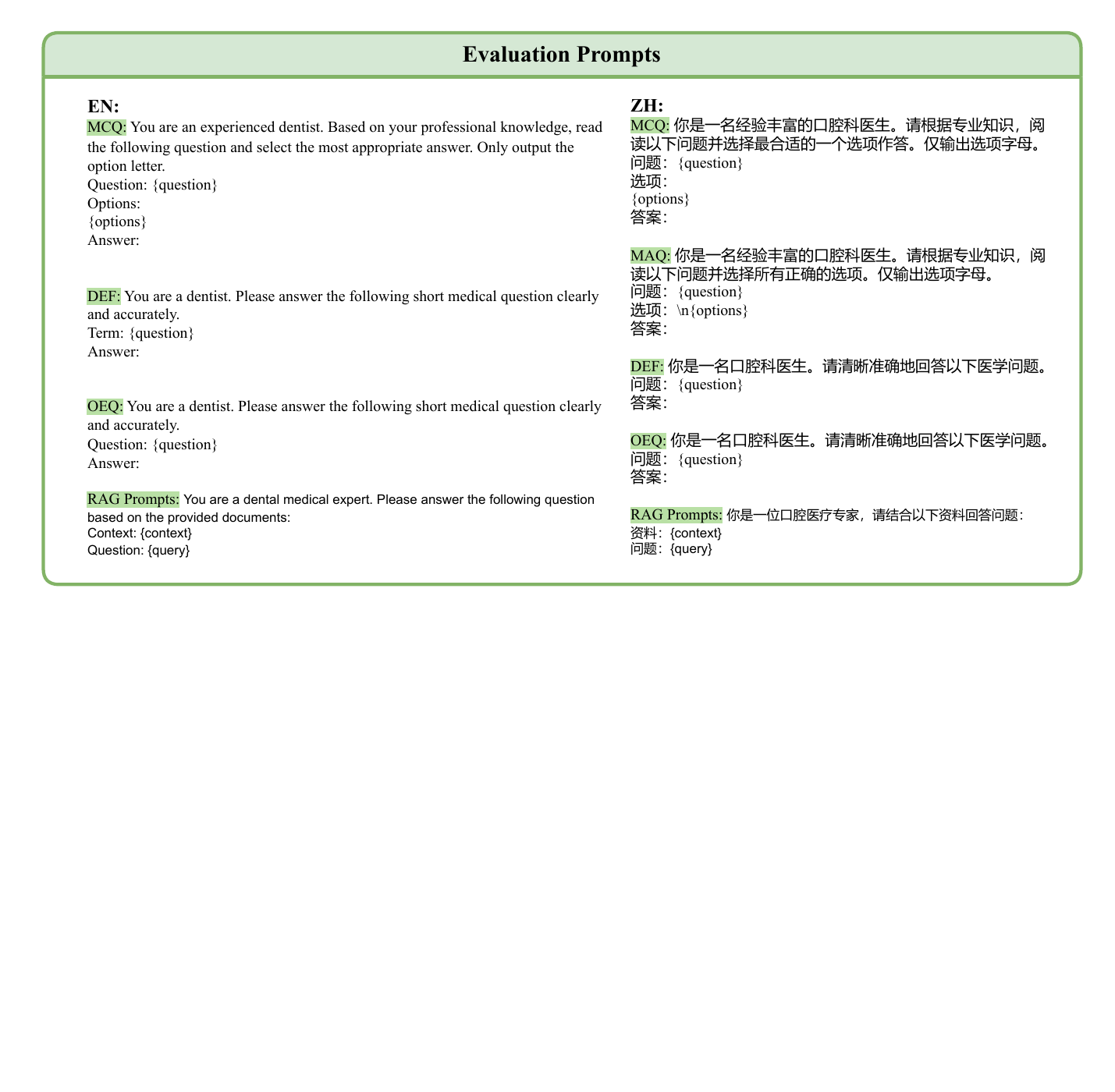}
    \caption{Evaluation and RAG Prompts}
    \label{fig:evalRAG}
    \vspace{-1em}
\end{figure*}

Fig.~\ref{fig:evalRAG} shows the prompt formats used to evaluate different question types in zero-shot settings and the prompt format with RAG.

\section{Corpus Construction Details}
\label{sec:corpus}

\subsection{MeSH Terms for PubMed Query}
\label{sec:mesh_terms}
\begin{figure*}
    \centering
    \includegraphics[width=1\linewidth]{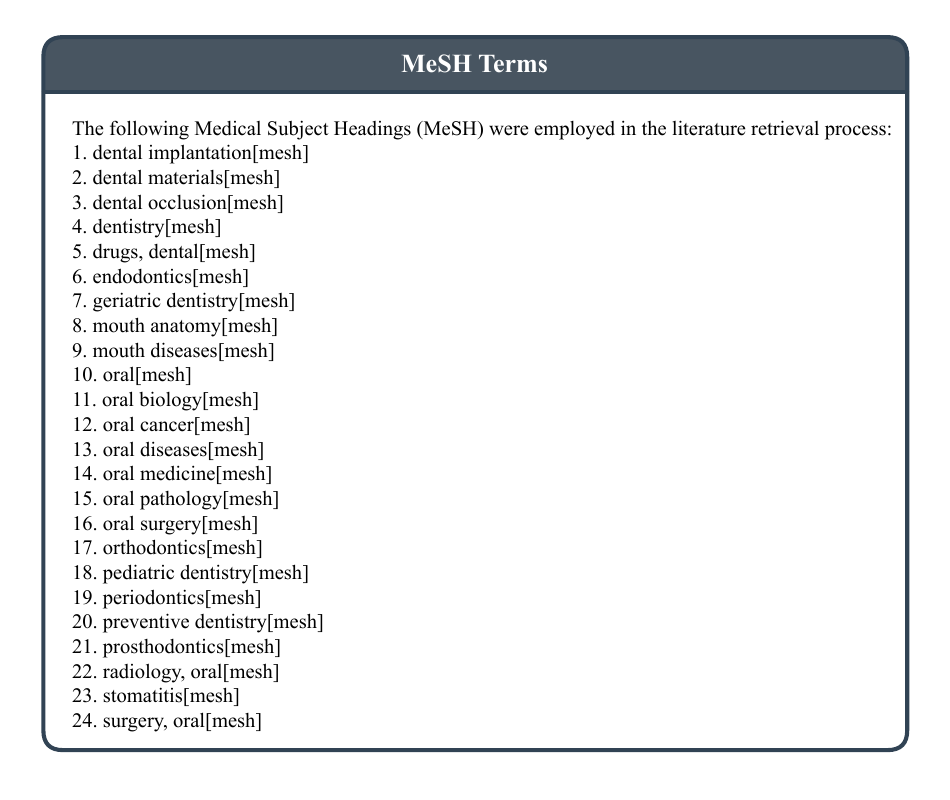}
    \caption{MeSH terms}
    \label{fig:mesh_terms}
    \vspace{-1em}
\end{figure*}

Fig.~\ref{fig:mesh_terms} is the list of 28 MeSH terms used to retrieve relevant dental articles from PubMed.

\label{sec:MeSH terms}

\section{Dataset Statistics and Visualizations}
\label{sec:stats}

\subsection{Distribution by Task and Subfield}
\label{sec:distribution}

\begin{figure*}
    \centering
    \includegraphics[width=1\linewidth]{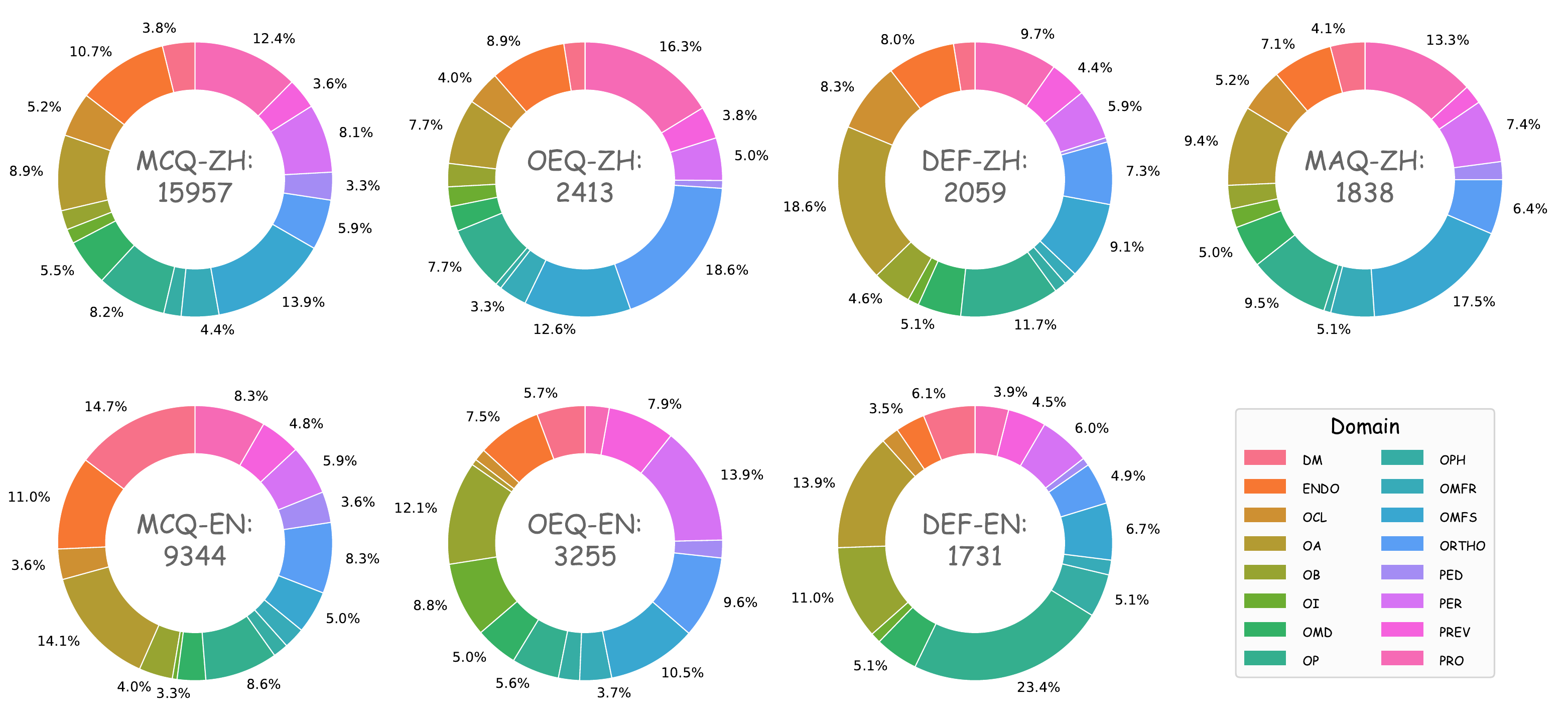}
    \caption{Distribution by Task and Subfield}
    \label{fig:Distribution}
    \vspace{-1em}
\end{figure*}

Fig.~\ref{fig:Distribution} shows the distribution of DentalQA by task and subfield.

\subsection{Answer Properties and Input Lengths}
\label{sec:qa_properties}

Fig.~\ref{fig:evalRAG} shows the prompt formats used to evaluate different question types in zero-shot settings and the prompt format with RAG.

\begin{table*}[!ht]
    \centering
    \caption{The average length of questions and answers in OEQ and DEF}
    \begin{tabular}{c|ccccc}
    \hline
        Question Format & Content & Mean & Median & Min & Max \\ \hline
        OEQ-EN & answer & 331.44  & 263 & 58 & 1941 \\ 
        OEQ-EN & question & 147.90  & 108 & 20 & 1498 \\ 
        OEQ-ZH & answer & 182.88  & 149 & 7 & 1321 \\ 
        OEQ-ZH & question & 25.21  & 16 & 6 & 326 \\ 
        DEF-EN & answer & 312.10  & 193 & 46 & 5249 \\ 
        DEF-EN & question & 77.84  & 75 & 24 & 234 \\ 
        DEF-ZH & answer & 59.79  & 52 & 2 & 254 \\ 
        DEF-ZH & question & 22.63  & 22 & 6 & 64 \\  \hline
    \end{tabular}
\end{table*}

\subsection{Supplementary Performance Figures}
\label{sec:perf_figures}
\begin{figure*}
    \centering
    \includegraphics[width=1\linewidth]{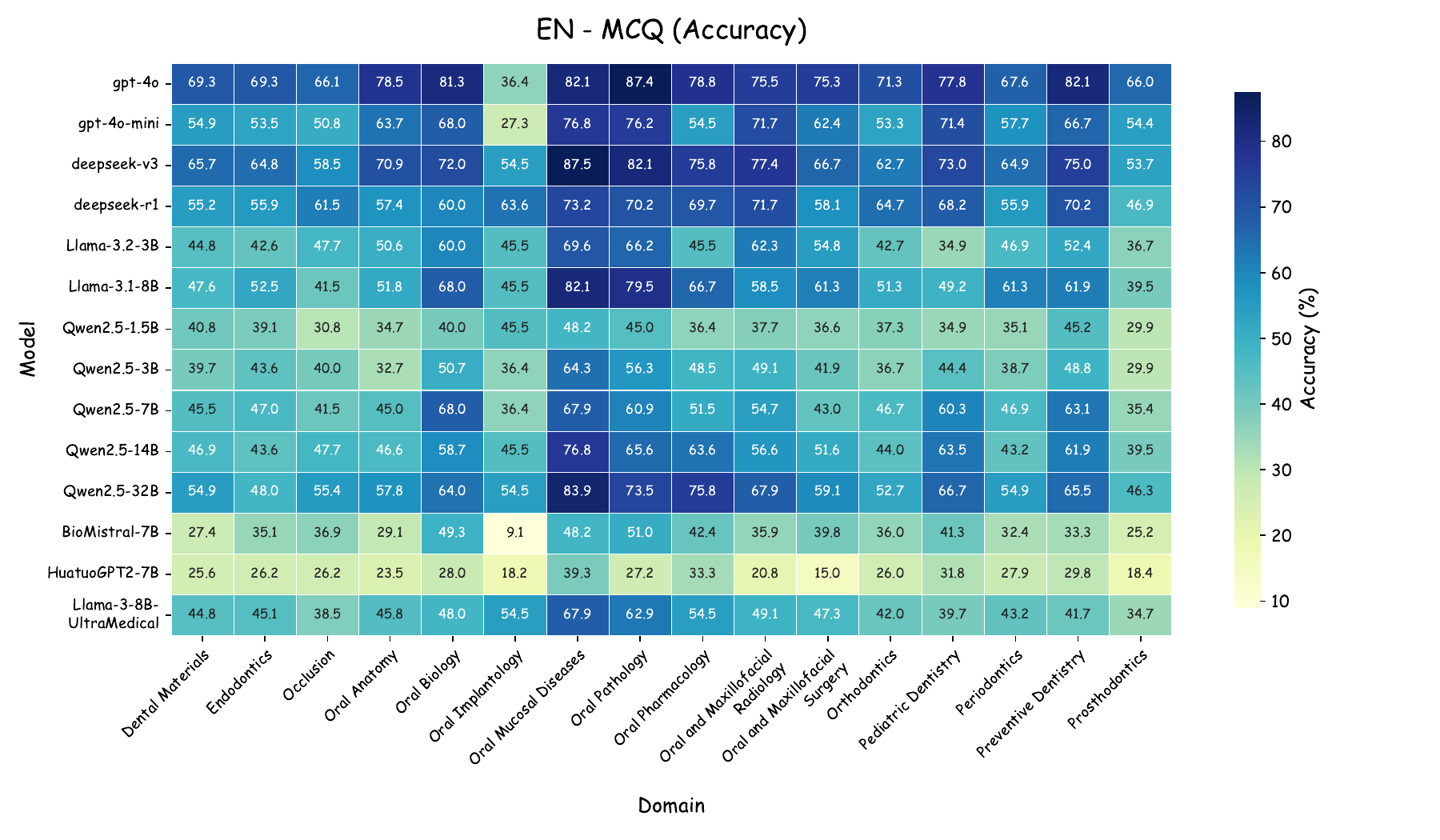}
    \caption{MCQ-EN-Accuracy}
    \label{fig:MCQ-EN-Accuracy-Distribution}
    \vspace{-1em}
\end{figure*}

\begin{figure*}
    \centering
    \includegraphics[width=1\linewidth]{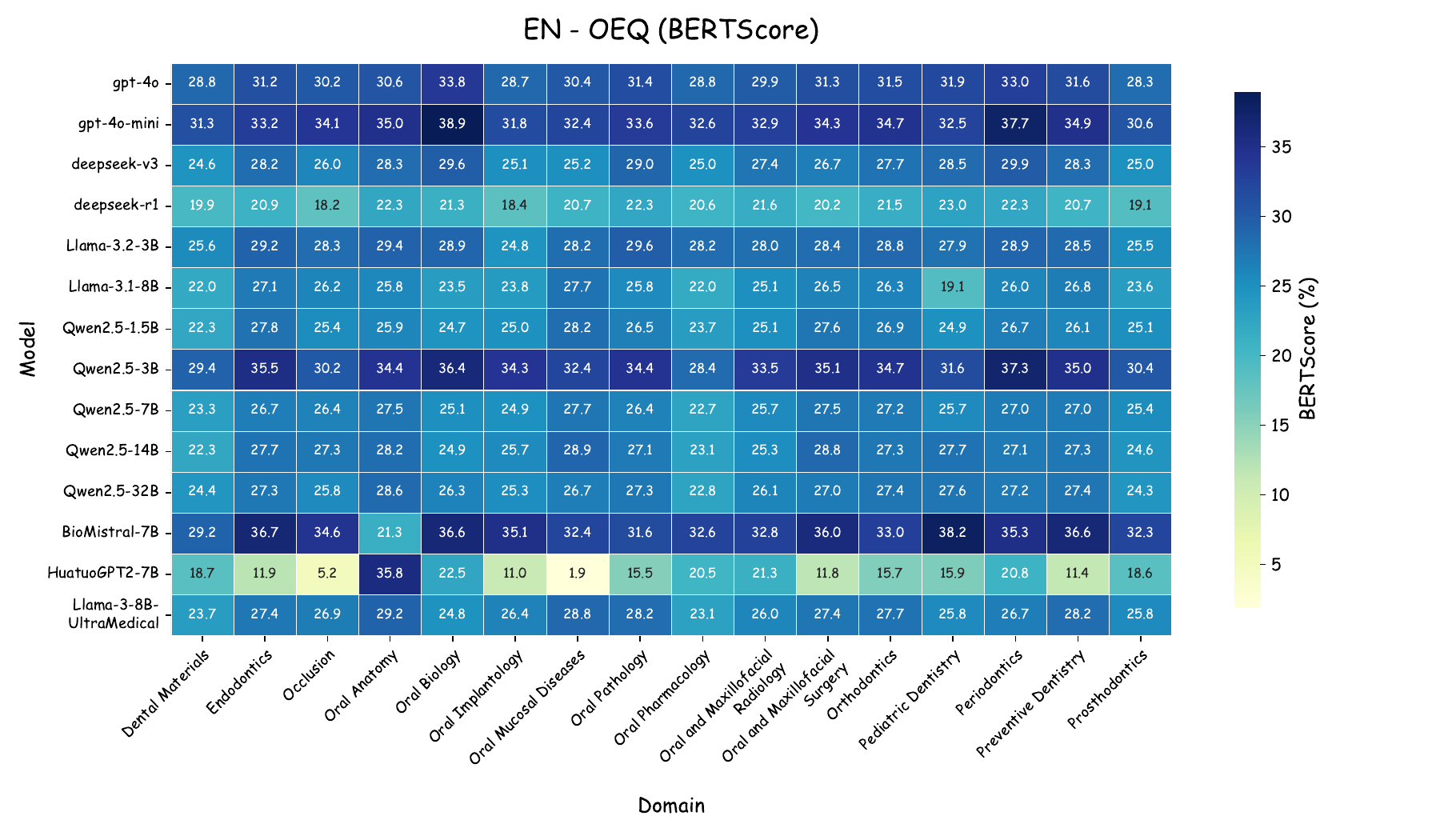}
    \caption{OEQ-EN-BERTScore}
    \label{fig:OEQ-EN-BERTScore-Distribution}
    \vspace{-1em}
\end{figure*}

\begin{figure*}
    \centering
    \includegraphics[width=1\linewidth]{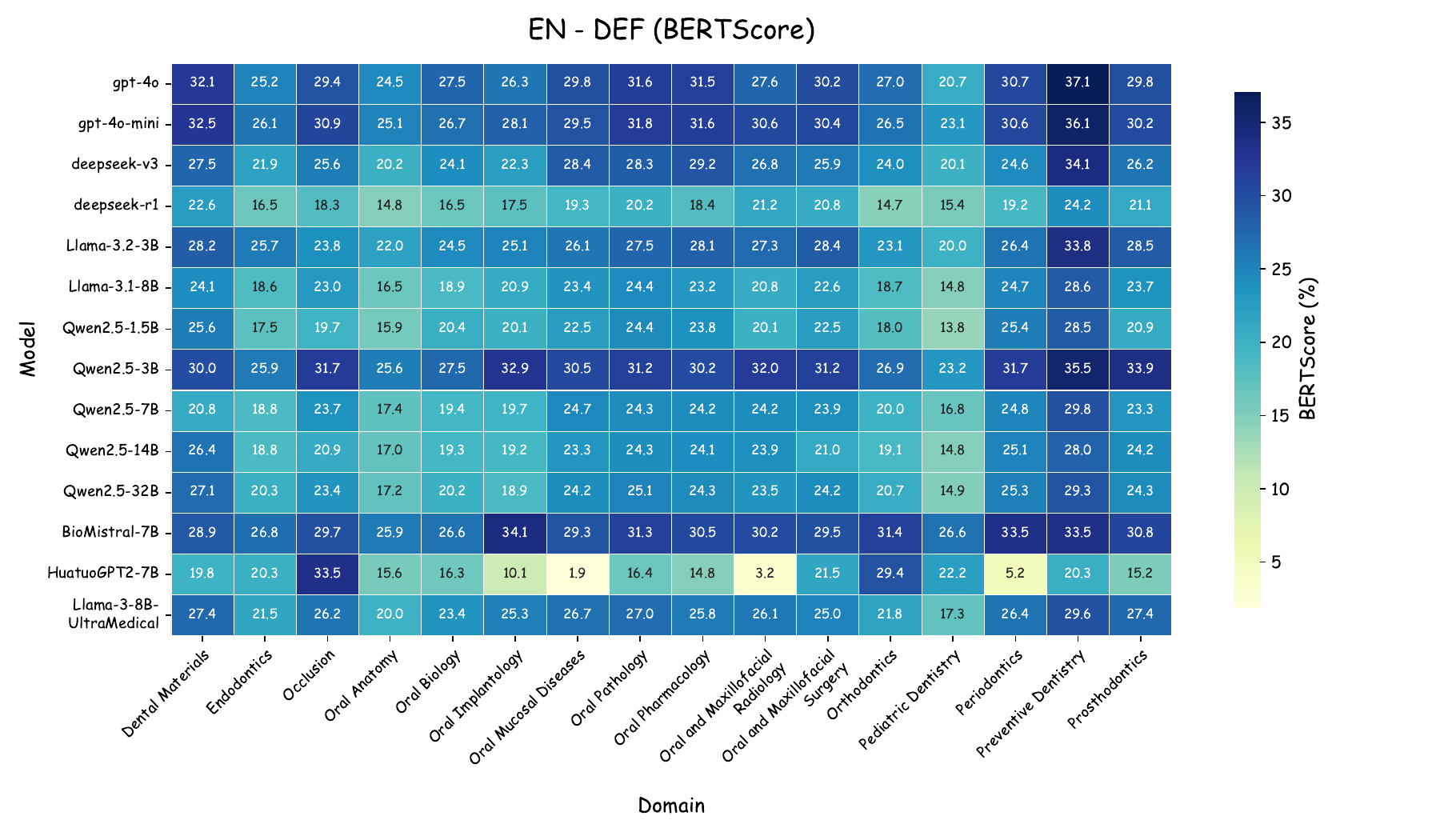}
    \caption{DEF-EN-BERTScore}
    \label{fig:DEF-EN-BERTScore-Distribution}
    \vspace{-1em}
\end{figure*}

\begin{figure*}
    \centering
    \includegraphics[width=1\linewidth]{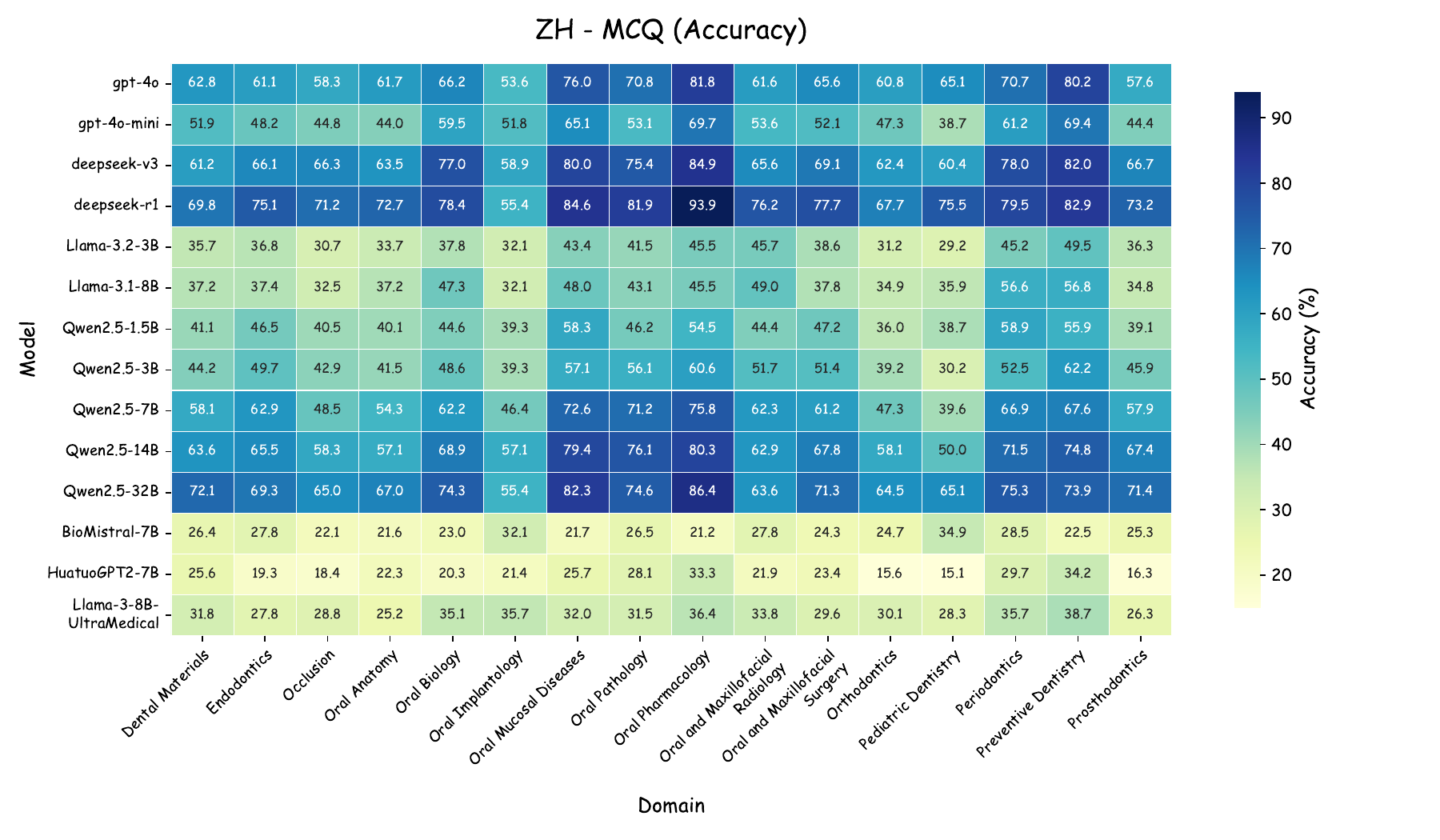}
    \caption{MCQ-ZH-Accuracy}
    \label{fig:MCQ-ZH-Accuracy-Distribution}
    \vspace{-1em}
\end{figure*}

\begin{figure*}
    \centering
    \includegraphics[width=1\linewidth]{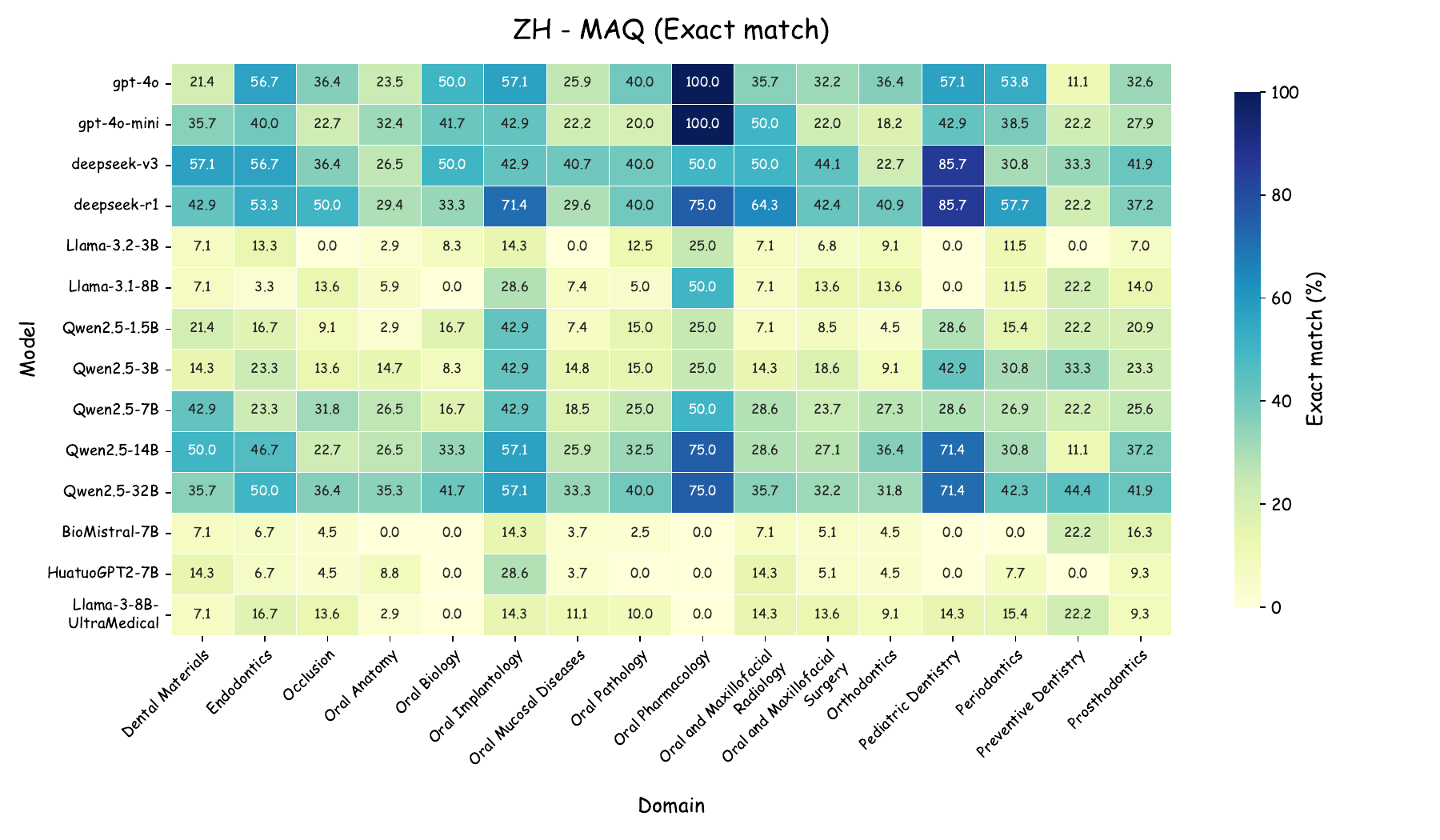}
    \caption{MAQ-ZH-Accuracy}
    \label{fig:MAQ-ZH-Accuracy-Distribution}
    \vspace{-1em}
\end{figure*}

\begin{figure*}
    \centering
    \includegraphics[width=1\linewidth]{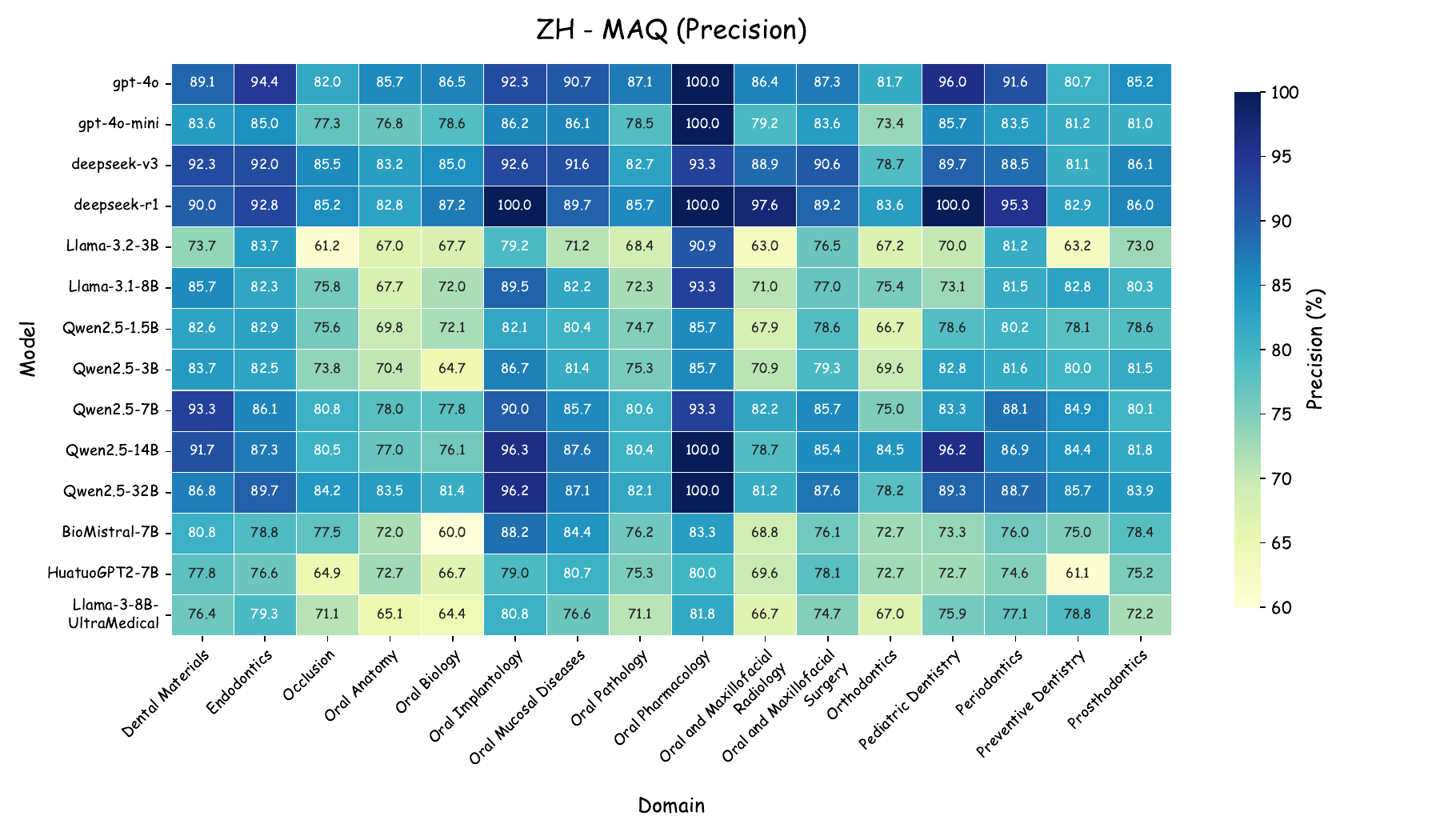}
    \caption{MAQ-ZH-Precision}
    \label{fig:MAQ-ZH-Precision-Distribution}
    \vspace{-1em}
\end{figure*}

\begin{figure*}
    \centering
    \includegraphics[width=1\linewidth]{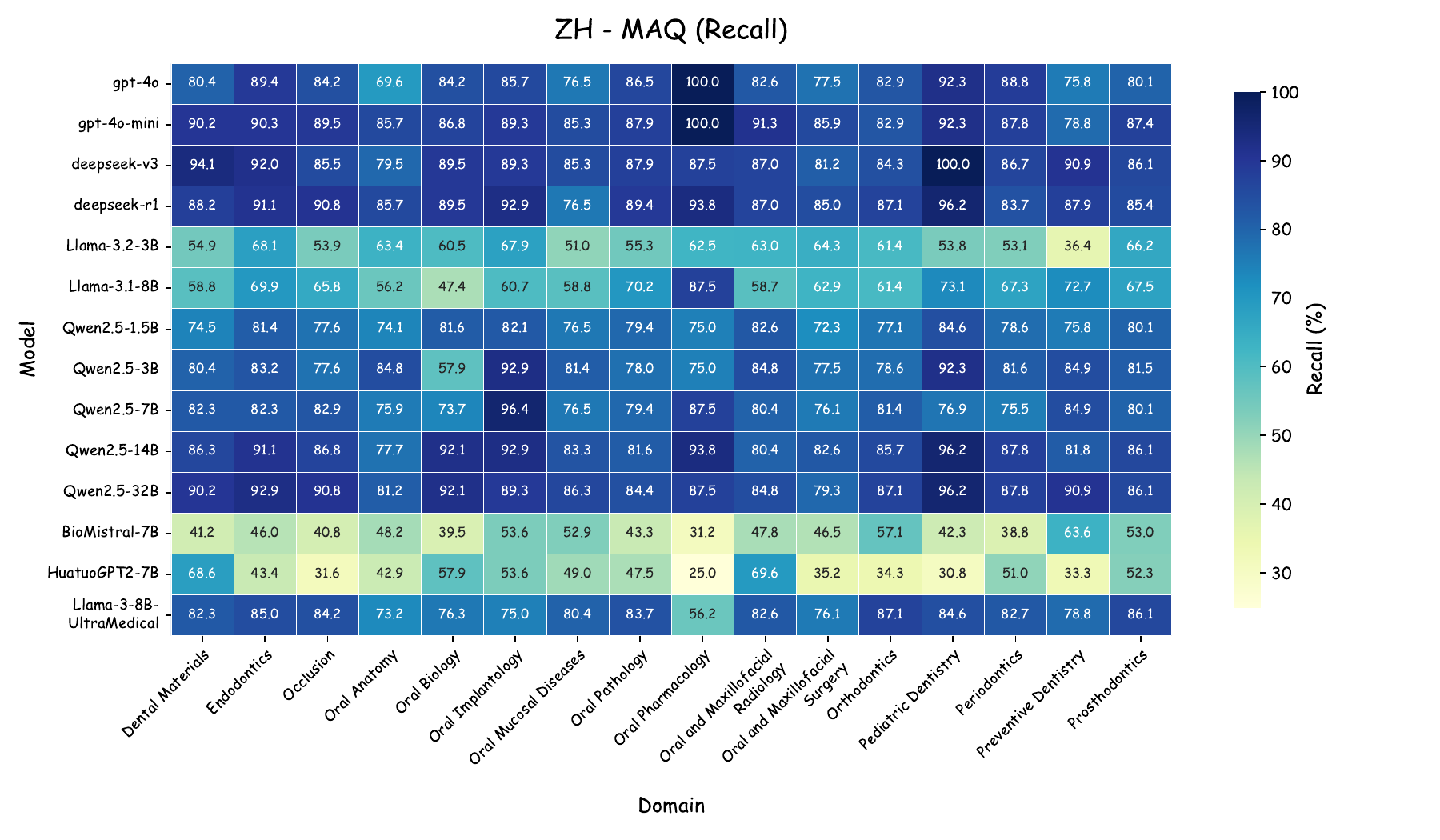}
    \caption{MAQ-ZH-Recall}
    \label{fig:MAQ-ZH-Recall-Distribution}
    \vspace{-1em}
\end{figure*}

\begin{figure*}
    \centering
    \includegraphics[width=1\linewidth]{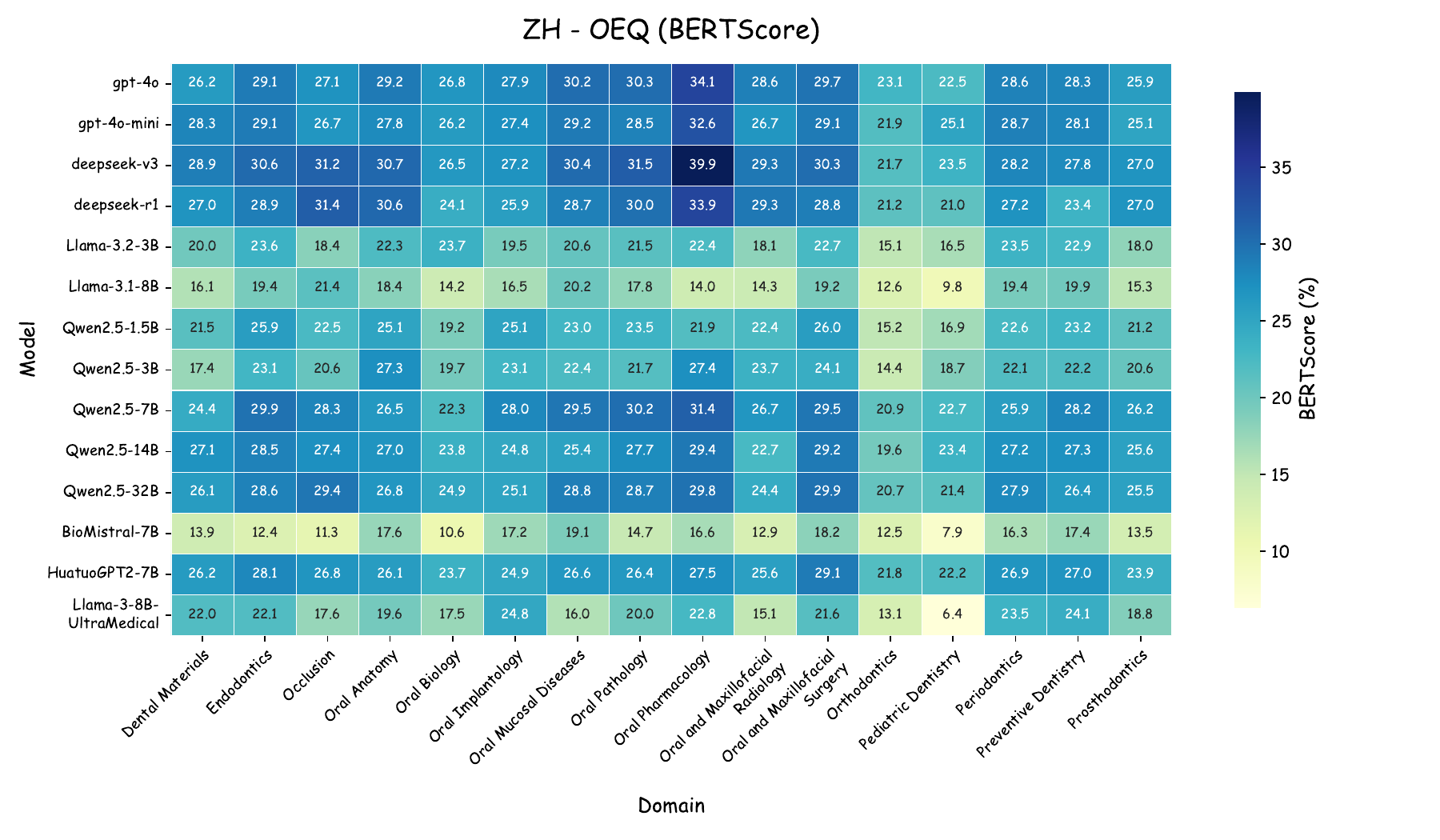}
    \caption{OEQ-ZH-BERTScore}
    \label{fig:OEQ-ZH-BERTScore-Distribution}
    \vspace{-1em}
\end{figure*}

\begin{figure*}
    \centering
    \includegraphics[width=1\linewidth]{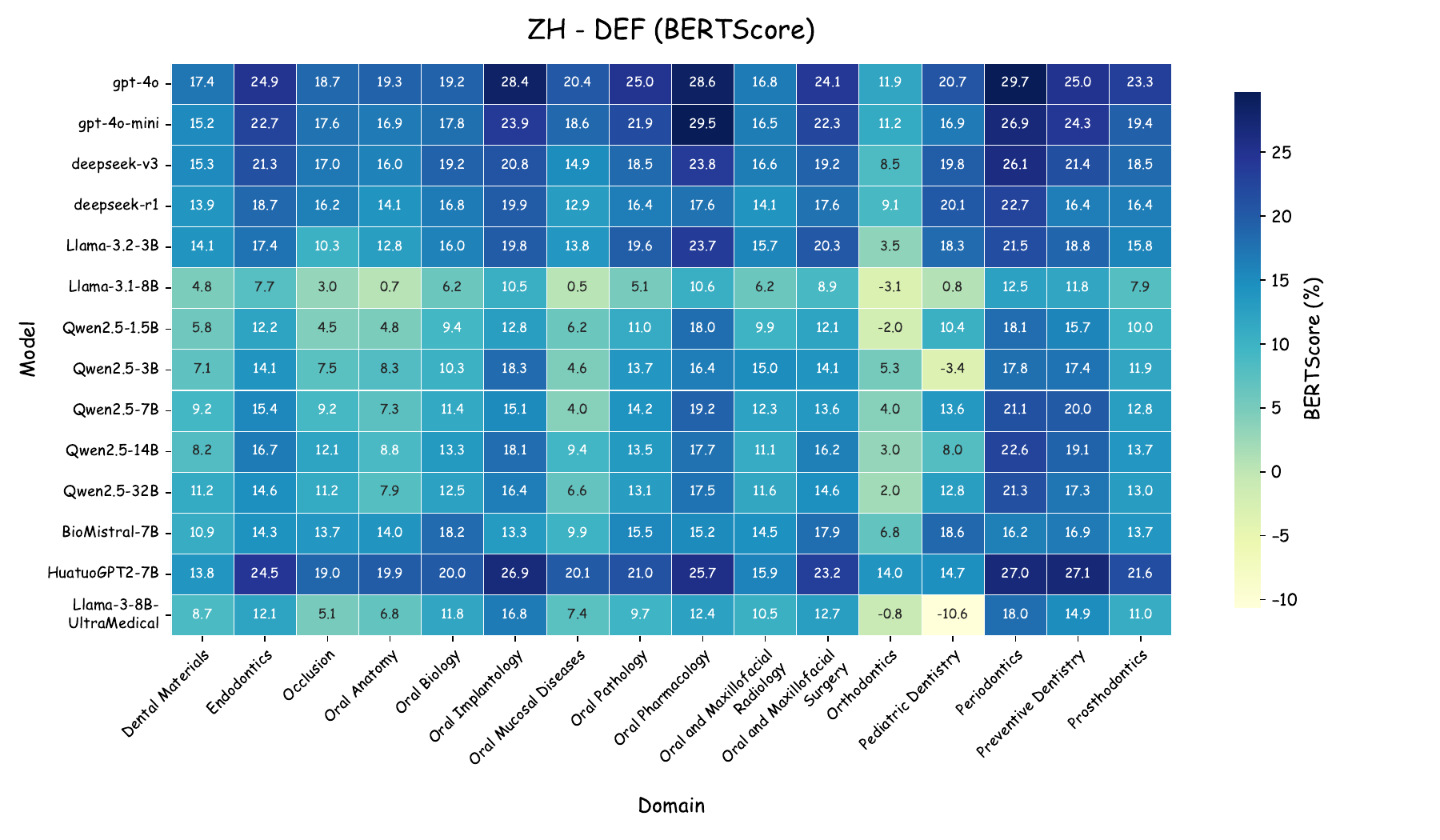}
    \caption{DEF-ZH-BERTScore}
    \label{fig:DEF-ZH-BERTScore-Distribution}
    \vspace{-1em}
\end{figure*}

Extended plots (\ref{fig:MCQ-EN-Accuracy-Distribution}, \ref{fig:OEQ-EN-BERTScore-Distribution}, \ref{fig:DEF-EN-BERTScore-Distribution}, \ref{fig:MCQ-ZH-Accuracy-Distribution}, \ref{fig:MAQ-ZH-Accuracy-Distribution}, \ref{fig:MAQ-ZH-Precision-Distribution}, \ref{fig:MAQ-ZH-Recall-Distribution}, \ref{fig:OEQ-ZH-BERTScore-Distribution}, \ref{fig:DEF-ZH-BERTScore-Distribution}) complementing Section~\ref{tab:oralkb_bench}, including per-model and per-task visual comparisons.

\begin{figure*}
    \centering
    \includegraphics[width=1\linewidth]{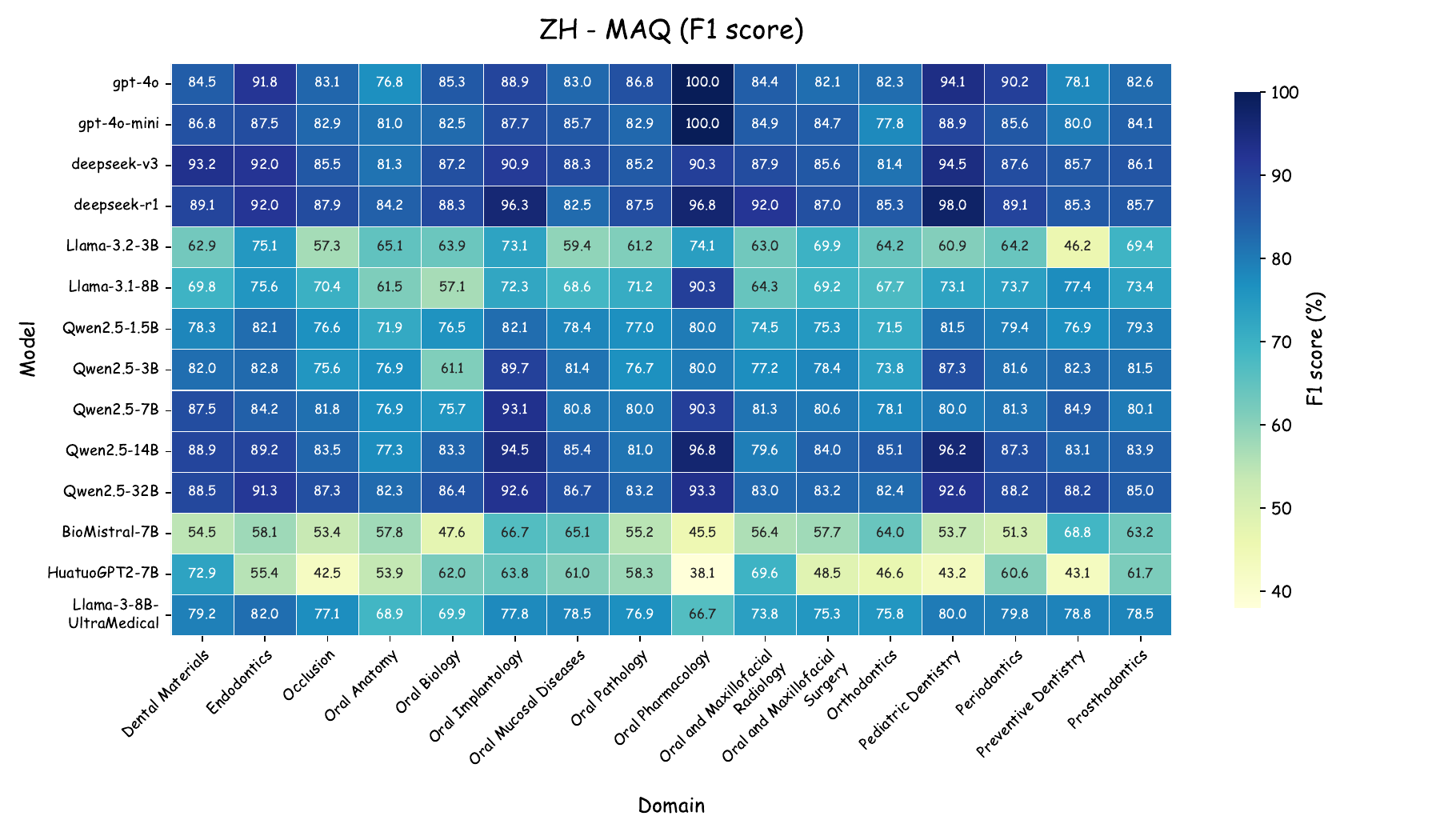}
    \caption{MAQ-ZH-F1}
    \label{fig:MAQ-ZH-F1-Distribution}
    \vspace{-1em}
\end{figure*}

\end{document}